\documentclass{article}
\usepackage{microtype}
\usepackage{graphicx}
\usepackage{booktabs}
\usepackage{hyperref}
\usepackage[preprint]{icml2026}
\hypersetup{pdfsubject={}}
\usepackage{amsmath,amssymb}
\usepackage{array}
\usepackage{longtable}
\usepackage{enumitem}
\setlist{nosep, leftmargin=*, after=\vspace{2pt}}

\newcolumntype{L}[1]{>{\raggedright\arraybackslash}p{#1}}

\icmltitlerunning{Testing Latent Intermediate Values with Minimal Pairs}

\begin{document}

\twocolumn[
  \icmltitle{Copied, Mixed or Computed? Testing Latent Intermediate Values with Minimal Pairs}
  \begin{icmlauthorlist}
    \icmlauthor{Aur Shalev Merin}{ind}
  \end{icmlauthorlist}
  \icmlaffiliation{ind}{Independent researcher}
  \icmlcorrespondingauthor{Aur Shalev Merin}{aur.merin@gmail.com}
  \icmlkeywords{probing, latent reasoning, looped transformers, interpretability}
  \vskip 0.3in
]

\printAffiliationsAndNotice{}

\begin{abstract}
Probing studies of latent reasoning read a model's state where the answer begins, and take a successful probe as evidence that an intermediate value was computed and a failed one as evidence that it was not. The second fails even for a digit copied from the question, and the first is ambiguous: using prompts that differ in a single token, we show that such a digit is present at that position in two looped and two standard language models, while a probe that sees one prompt at a time reads it at no more than 0.52 even after 45,000 training questions, and every token we varied leaves such a trace, so a probe for a computed value may be reading the inputs. We give a protocol that tests a candidate computed value against copies of the inputs, against generic mixing of them with an operation swap whose bar is set on synthetic mixtures, and for use with a clamp that controls for the copy. We check each test where the answer is known. The clamp detects values the model is about to write or reads four tokens later; in two-hop recall in a pretrained model it finds the unwritten bridge entity where whole-state patches show it is used, and stays silent at a later token where the entity is readable (0.92--1.00 from layer 12) but, as whole-state patches show, unused; and in small transformers trained from scratch it detects an unwritten value used eight tokens later. Our first, accuracy-based swap failed its positive control, so we replaced it with a version based on shares of variance that passed two fresh controls before we applied it, though in the small transformers it fires wherever the value is present, used or not, and in one of 270 null worlds. In the looped Ouro-1.4B the protocol finds the unwritten intermediate's sign following the operation beyond copies, the answer, the carry and borrow, and generic mixing of the operands, though narrowly and as a small part of the state, which cannot be told apart from the answer with a copied number added back, and no running total shows before that number is read; its use at the position tested is unresolved: a clamp moves the answer in 4\% of pairs, against 0.3\% for a random one. For a published claim that language models compute sub-answers only while writing a chain of thought, a standard probe reads the sub-answer at up to 0.58 at the last prompt token before a direct answer, and a test against copies alone and a clamp that leaves the copy in place both suggest an early sub-answer. In correctly answered direct-answer problems, replacing the whole state at all five positions between the sub-answer's equation and the token before the answer, at every layer, moves the answer to the one a carried sub-answer predicts in 0.2--0.3\% of pairs, against 88--90\% for an operand's own token, so the sub-answer is not carried on its own past its equation, consistent with the claim. In pairs whose answers differ it gives the partner's answer in 8\% (above zero, below our bar), so a weak carry of the answer before it is written is not ruled out; a post hoc map places most of it at ``='', the last of the five. A probe could not have shown which positions are used; with a chain the protocol cannot tell.
\end{abstract}

\section{Introduction}\label{sec:intro}

Looped transformers apply one block of layers repeatedly \citep{geiping2025,zhu2025ouro}, and one hope is that the extra loops work as a hidden chain of thought; a related question is whether standard models compute sub-answers before they write a chain at all \citep{kudo2026}. Accuracy alone cannot settle either. Answering directly after the question, Ouro-1.4B's GSM8K accuracy rises from 6.5\% after one loop to 23\% after four, but almost all of the gain comes at loop 2, where Ouro also improves on benchmarks that involve no arithmetic (Appendix~\ref{app:acc}). To see whether intermediate values are computed, one has to look inside.

Probing studies of latent reasoning read the state at the position where the answer begins, which we call the summary position, and take a successful probe as evidence that a value was computed and a failed one as evidence that it was not \citep{lu2025,yu2024,kudo2026}. The second fails even for the simplest value there is, a digit copied from the question, and the first is ambiguous for the same reason. Our main contribution is a protocol for computed values whose tests are each checked where the answer is known:
\begin{itemize}
\item \textbf{The problem} (Section~\ref{sec:probes}). Minimal pairs, prompts that differ in a single token, show the copied digit at the last prompt token of two looped models (Ouro-1.4B and 2.6B) and two standard ones (Qwen2.5-1.5B and 7B), 0.79--0.88 on average in three models and 0.92--0.97 across Ouro-1.4B's states. A probe that sees one prompt at a time, as probing studies use, reads it at 0.12--0.27 with tens of training templates and at most 0.52 after 45,000 training questions. Every token we varied leaves such a trace, so a probe that reads a ``computed'' value may be reading the inputs.
\item \textbf{The protocol} (Section~\ref{sec:protocol}). A candidate value is tested against copies of the inputs, against generic mixing with an operation swap whose bar is set on calibrated synthetic mixtures, and for use with a clamp that controls for the copy. Each step rejects something that a weaker test accepts in our data (Table~\ref{tab:steps}).
\item \textbf{Validation} (Section~\ref{sec:control}). In pre-registered positive controls the copy-controlled clamp detects a value the model is about to write (+0.61 over a random clamp) and one it reads four tokens after writing it (+0.81). Our first, accuracy-based operation swap failed its control, because a code of the operand pair makes every function of the pair readable; a swap based on shares of variance (the share swap), frozen before testing, passed two fresh controls in two model families, with no false positive in 420 null worlds (later, 1 in 270 on another). In two-hop recall in Qwen2.5-7B the clamp finds the unwritten bridge entity where whole-state patches show it is used (+0.79 over a random subspace) and stays silent at the last token, where the entity reads at 0.92--1.00 but the whole state changes nothing either. In small transformers trained from scratch it also detects an unwritten value used eight tokens later, while the share swap fires wherever that value is present.
\item \textbf{Two applications.} In Ouro-1.4B (Section~\ref{sec:ouro}) the sign of the unwritten intermediate $a+b$ follows the operation just before the answer's last digit, beyond the copies, the answer, the carry and borrow, and a bar set by generic mixing, though narrowly and as under 1\% of the remaining variance; it cannot be told apart from the answer with $c$ added back, and its use is unresolved (a clamp moves the digit in 4\% of pairs, random 0.3\%). For the claim of \citet{kudo2026} that sub-answers are computed only while the chain is written (Section~\ref{sec:kudo}), weaker tests suggest an early sub-answer, and the protocol does not confirm one. Without a chain, in correctly answered problems, replacing the whole state at the five positions between the sub-answer's equation and the token before the answer, at every layer, moves the answer to the one a carried sub-answer predicts in 0.2--0.3\% of pairs, so the sub-answer is not carried on its own; with different answers it gives the partner's answer in 8\% (UNCLEAR: above zero, below our bar), mostly from the ``='' just before the answer position (post hoc). This is consistent with their conclusion for the sub-answer carried on its own; for $b$ together with $k$, or the answer, a weak early carry is not ruled out. Their probe, which reads the sub-answer at 0.58 at the last prompt token, where the answer uses nothing, could not show which positions are used. With a chain the protocol cannot tell.
\end{itemize}
Most studies were pre-registered, and the text marks the analyses that were not; Appendix~\ref{app:status} gives the status of every analysis.

\section{Setup}\label{sec:setup}

Ouro-1.4B and 2.6B \citep{zhu2025ouro} run their layers up to four times; we read the state in the middle of loop 1 and at the end of each loop, in fp32 (on check prompts our forward pass matches Ouro's own, with a largest logit difference of 0). Qwen2.5-1.5B and 7B \citep{qwen2025} are standard transformers, read after layers 7, 14, 21 and 28 (and 4 in the 1.5B). All four give each digit its own token. Every prompt in Sections~\ref{sec:probes} and~\ref{sec:ouro} starts with the first four GSM8K training problems \citep{cobbe2021}, each followed directly by its final answer, then the test question and ``Answer:''.

The \emph{standard probe} standardizes the state, projects it onto 256 principal components fitted on the training folds, and fits a multinomial logistic regression whose regularization is chosen by inner cross-validation. It is scored on 5 held-out folds grouped by template, so it must generalize to unseen kinds of problem. The \emph{minimal-pair probe} is the same probe on states from which each family's mean has been subtracted. Labels are units digits. We probe P0, the last prompt token, and, in Section~\ref{sec:ouro}, P1: with the gold two-digit answer fed in, the tens-digit token, after which the units digit is predicted.

\section{Why a probe at the summary position is ambiguous}\label{sec:probes}

\paragraph{A copied digit is present but hard to read.}\label{sec:presence} At the last prompt token a standard probe does not read a digit copied from the question. In a pre-registered study its sanity check, the last digit of the largest number in the question, scored 40.8\%, exactly the rate of always guessing the most common digit; by position, a copied digit is read perfectly on its own token and at chance from three tokens on (Appendix~\ref{sec:mp}). We therefore took one question per template and one of its multi-digit numbers, and made ten copies that differ only in that number's last digit: 137 families from 77 distinct questions. The copies are identical token for token except one, so any difference in a later state is caused by that digit. Counterfactual minimal pairs are a standard interpretability control \citep{stolfo2023,vig2020investigating}; subtracting a family's mean removes what the family shares, the contrast that difference-in-means probing uses to find a direction \citep{marks2023}.

\begin{figure*}[t]
\centering
\includegraphics[width=0.9\linewidth]{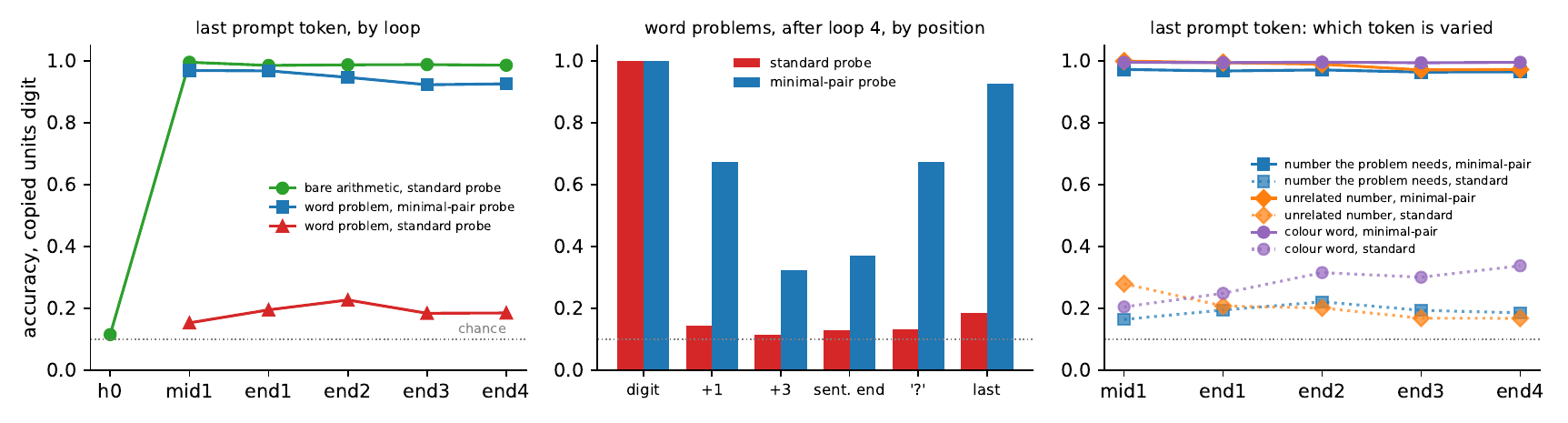}
\caption{Reading a copied token in Ouro-1.4B (chance 0.10). Left: the last prompt token, by loop. Middle: word problems after loop 4, by position; ``+k'' is k tokens after the digit. Right: the last prompt token, for a number the problem needs, a number in an added, irrelevant sentence, and a colour word in that sentence. Solid lines: the minimal-pair probe; dotted: the standard probe (right panel: means over two insertion points, so the needed number differs slightly from the per-state values in Appendix~\ref{sec:mp}).}
\label{fig:readout}
\end{figure*}

\begin{table*}[t]
\centering
\caption{The copied digit at the last prompt token (chance 0.10). Upper rows: 137 families from 77 questions; ranges over Ouro's states and Qwen's layers 7--28, with the mean in parentheses. Lower rows: 1,000 held-out GSM8K test questions; minimal pairs on 100--150 of them; single-prompt probes best over states and over PCA-256 and full-state logistic regression. Pre-registered verdicts are in Appendix~\ref{sec:many}. Ouro-1.4B's family study was exploratory. The best single-prompt state is picked on the test set, which favours the single-prompt probe. $^\ast$Qwen2.5-7B's last layer; at most 0.29 at its other layers.}
\label{tab:presence}
\small
\setlength{\tabcolsep}{4pt}
\begin{tabular}{@{}lcccc@{}}
\toprule
 & Ouro-1.4B (looped) & Ouro-2.6B (looped) & Qwen2.5-1.5B & Qwen2.5-7B \\
\midrule
minimal-pair probe & 0.92--0.97 & 0.72--0.92 (0.79) & 0.75--0.98 (0.88) & 0.58--0.99 (0.85) \\
standard probe & 0.15--0.23 & 0.14--0.17 & 0.12--0.14 & 0.13--0.27 \\
structured decoder & 0.17--0.21 & 0.14--0.17 & 0.12--0.14 & 0.14--0.26 \\
\midrule
minimal pairs on test questions & 0.96 & 0.79 & 0.91 & 0.86 \\
single prompt, 5,249 training questions & 0.36 & 0.22 & 0.19 & 0.43 \\
single prompt, 45,249 training questions & -- & 0.31 & -- & 0.52$^\ast$ \\
\bottomrule
\end{tabular}
\end{table*}

The minimal-pair probe reads the digit on templates it never saw, 0.79--0.88 on average in three models and 0.92--0.97 across Ouro-1.4B's states, while the standard probe reads it at 0.12--0.27 (Table~\ref{tab:presence}, Figure~\ref{fig:readout}). A structured decoder, the best single-prompt decoder we could build (a mass-mean probe from the minimal-pair digit means with a shrunk covariance of the training prompts), does no better (Appendix~\ref{sec:many}). The code is shared across questions: in Ouro-1.4B a probe trained on needed numbers reads a number in an added, irrelevant sentence at 0.78--0.97, and a minimal-pair probe of a colour word in that sentence reads it at 0.99--1.00. It is not a magnitude direction \citep{zhuf2024}, since the probe's errors are no closer to the true digit than random guesses, which fits per-digit codes \citep{levy2025}. The minimal-pair probe needs all ten variants of a question, so it tests presence; it is not a decoder for a new prompt.

\paragraph{More questions do not close the gap.}\label{sec:more} In pre-registered studies we gave each model one prompt for each of 6,249 GSM8K training questions, with the units digit of the last number of two or more digits set at random, held out a fixed 1,000, and trained probes on nested subsets of the rest (Figure~\ref{fig:curve}); for the larger models we added 40,000 Orca-Math questions \citep{mitra2024orcamath}, after removing any that reword a test question. At 5,249 training questions the best single-prompt probe reads 0.19 in Qwen2.5-1.5B and 0.36 in Ouro-1.4B; at 45,249 it reads 0.31 in Ouro-2.6B and 0.52 at Qwen2.5-7B's last layer (at most 0.29 at its other layers), against 0.79--0.96 for minimal pairs on the same test questions. The last doubling adds under 0.01 on the GSM8K test questions but still 0.014--0.018 on Orca-Math ones, so data closer to the test questions may read more. The claim is that presence and single-prompt readability come apart by a wide margin at the data sizes tested, not that the digit cannot be read.

\paragraph{Why it is hard to read.}\label{sec:small} The digit's directions are no more crowded than those of a random word swapped in for it or of noise added to its embedding; its change is simply small, and the standard probe reads it about as poorly as the noise (pre-registered; Appendix~\ref{sec:swamp}).

\paragraph{Both kinds of result are ambiguous.} A failed probe at a summary position is weak evidence that a value is absent: a digit that is demonstrably present reads below 0.30 in most states. This concerns probes that must generalize to new questions; on prompts that differ only in their numbers, such as bare additions, a standard probe reads both operands at 0.96--1.00 (Appendix~\ref{sec:mp}). A successful probe is weak evidence of computation: every token we varied, an irrelevant number and a colour word included, leaves a trace at the same position, so a probe that reads a ``computed'' value may be reading the inputs it is a function of. In small transformers without varied context a computed value shows no such gap (post hoc, Appendix~\ref{app:v230}); computed values in language models are untested. The protocol addresses the second problem.

\section{A protocol for computed values}\label{sec:protocol}

Within a minimal-pair family one digit varies, so any computed value is a fixed function of it. The protocol therefore varies the operands together in factorial families, with everything else fixed, and asks five questions of a candidate value such as $(a+b) \bmod 5$:
\begin{enumerate}
\item \textbf{Is it more than copies?} If the state holds $f(a)+g(b)$, a linear probe on it does no better than the same probe on one-hot codes of the digits, since any function of a digit is linear in its one-hot code. We call that probe's accuracy the \emph{copy ceiling}, a conditional probe with the inputs as its baseline \citep{hewitt2021conditional}, and score every label as its excess over its own ceiling. A prime modulus keeps other combinations from sharing the target's parity. Excess shows only that the operands were COMBINED.
\item \textbf{Does it follow the operation?} A generic mixture makes many combinations readable, $a+b$ and $a-b$ alike. Problems therefore come in twin families that differ only in the operation. After regressing the copies (and, where it confounds, the answer) out of the family-centred states, we take the share of the remaining variance that each label's class means explain on held-out families. With $D_+$ the share of $a+b$ minus that of $a-b$ on plus problems, and $D_-$ the reverse on minus problems, $I = (D_+ + D_-)/2$ cancels any fixed preference for one sign. FOLLOWS requires both $D$s positive, $I$'s interval above zero, and $I$ at least 0.30 of $N$, the needed label's share ($a+b$'s on plus problems and $a-b$'s on minus ones, averaged); DOES-NOT-FOLLOW, $I$'s upper bound below $0.10N$; otherwise UNCLEAR. We call this the share swap. An earlier version compared probe accuracies instead (the accuracy swap); it failed its positive control (Section~\ref{sec:control}).
\item \textbf{Is it more than mixing?} Synthetic states hold copies of the inputs and one generic nonlinear mixture (products of random codes for pairs of inputs, same-frequency products of periodic digit features, a random network, or a random code for each operand pair, with or without the operation), but no term for the candidate. The bar of 0.30 is 1.5 times the largest ratio among 420 such worlds of varied mixing strength (0.19), rounded up; none of them passes it, nor any of 420 fresh ones (Section~\ref{sec:control}).
\item \textbf{Does it rank first?} (Descriptive; not part of the decision.) Compare the target with every combination of the inputs, since periodic number features \citep{zhou2024fourier,kantamneni2025trigonometry} make some combinations easier to read than others.
\item \textbf{Is it used?} Replace the state's component in the candidate's subspace (spanned by family-centred class means, fitted on other families) with that of a partner problem, and read the output. Replacing the whole state is the positive control, and a random subspace the negative one. The subspace is projected off the copy's (and, before an answer, the answer's), since otherwise a clamp can act through the copied digit. USED needs a gain of at least 0.10 over the random clamp with the interval above zero; NOT-USED, an upper bound below 0.02.
\end{enumerate}
Results between a step's bars are UNCLEAR, and a test whose accuracy gate or positive control fails is UNSCORABLE. A computed value needs COMBINED and FOLLOWS, and a used one USED as well. Each step rejects something a weaker version accepts in this paper's data (Table~\ref{tab:steps}); Section~\ref{sec:control} tests each step where the answer is known.

\begin{table}[t]
\centering
\caption{What each step of the protocol rejects that a weaker test accepts, or finds that a weaker test misses, in this paper's data.}
\label{tab:steps}
\footnotesize
\setlength{\tabcolsep}{3pt}
\begin{tabular}{@{}L{0.22\linewidth}L{0.62\linewidth}l@{}}
\toprule
step & weaker test, and what it accepts & \S \\
\midrule
1. beyond copies & a standard probe: an irrelevant number and a colour word leave a trace at the summary position & \ref{sec:probes} \\
2. operation swap & excess over copies: every position in the Kudo task is combined (+0.12 to +0.38) & \ref{sec:kudo} \\
2. variance share & accuracy swap: random functions of the pair read 0.99--1.00, so a computed $b$ fails & \ref{sec:control} \\
3. calibrated bar & uncalibrated accuracy swap: generic mixing alone reaches $I = +0.14$, against Ouro's +0.157 & \ref{sec:ouro} \\
5. copy-controlled clamp & a clamp that keeps the copy moves the written $b$ in 30\% of pairs and the answer in 28--35\%; with the copy projected out, 2\% and 1--3\% & \ref{sec:kudo} \\
\bottomrule
\end{tabular}
\end{table}

\section{Validating the protocol}\label{sec:control}

A protocol that never says yes would tell us little, so pre-registered studies tested it where a computed value is certainly used. One uses two-hop recall and one small transformers trained from scratch; the others use the two-step problems of \citet{kudo2026}, $a = k \pm b$ and $b = d_1 \pm d_2$ with single-digit values, in a short chain format in which the model writes $b$ and then $a$ (``$b = 5$, $a = 6$''); Qwen2.5-7B writes $b$ correctly in 91\% of 12,000 prompts. The clamp controls were run after the applications in Sections~\ref{sec:ouro} and~\ref{sec:kudo} had found no USED value; the share swap was validated before it was applied (Table~\ref{tab:control}).

\begin{table}[t]
\centering
\caption{Each test where the answer is known (pre-registered). Power: the test says yes where the value is used or computed; false yes: where it is not. Clamps: gain over a random subspace, 1,000 pairs (whole-state patch 0.95 and 0.89 in Qwen; two-hop, 1,038 pairs, 0.85). Two-hop: the false-yes test is the last token, where the bridge entity is readable but unused. Small models: 6-layer transformers trained from scratch, value eight tokens before the answer (Appendix~\ref{app:v230}).}
\label{tab:control}
\footnotesize
\setlength{\tabcolsep}{2pt}
\begin{tabular}{@{}L{0.40\linewidth}L{0.29\linewidth}L{0.27\linewidth}@{}}
\toprule
test: setting & power & false yes \\
\midrule
clamp: Qwen-7B, $b$ written next & USED, +0.61 & random 0.00 \\[1pt]
clamp: Qwen-7B, $b$ read 4 tokens on & USED, +0.81 & random 0.00 \\[1pt]
clamp: Qwen-7B, two-hop bridge, never written & USED, +0.79 & NOT-USED, 0.00 (whole state 0.00 too) \\[1pt]
clamp: small models & USED 5/5 & none \\[1pt]
clamp, Ouro's form: at Ouro's signal & USED 5/5 to share 0.005 & -- \\[1pt]
clamp, Kudo's form (copy projected) & UNINFORMATIVE: USED 3/5 at full strength, 0/5 at 0.005; 5/5 when $\ge$55\% kept & -- \\[1pt]
accuracy swap: Qwen-7B, $b$ & fails & -- \\[1pt]
share swap: Qwen-7B $a$, Mistral $b$ & FOLLOWS ($I/N$ 0.65, 0.48; $N$ 0.11, 0.15) & 0/420 nulls \\[1pt]
share swap: small models & FOLLOWS 5/5 & unused positions; 1/270 nulls \\
\bottomrule
\end{tabular}
\end{table}

\paragraph{The clamp.} One token before $b$ is written, replacing $b$'s directions, with the copy projected out, makes the model write the partner's $b$ in 61\% of pairs, against none for a random subspace (USED). At $b$'s written digit, which the model reads four tokens later to compute $a = k \pm b$, it changes $a$ in 81\% (USED; 63\% where $a = 0 + b$ copies it). Both results are close to guaranteed, since $b$ is the next token or the token itself. At the end of $b$'s equation, a clamp of the copied digit's directions moves the written $b$ six tokens later in 96\% of pairs. Power for a value computed but never written is shown in two-hop recall and in the small models below.

\paragraph{The operation swap.} The accuracy-based swap failed: one token before $b$ is written, the operands are combined (+0.25) but $I = +0.017$, below its bar, so the protocol did not call $b$ computed. In an exploratory diagnosis the cause was saturation: there, four random functions of the operand pair read at 0.99--1.00, as $a+b$ and $a-b$ do, so accuracy has no room, while the share of residual variance still separates them. We therefore froze the share swap (Section~\ref{sec:protocol}), set its bar from 420 synthetic mixing worlds, and judged it only on controls it had never seen: on new problems in Qwen2.5-7B where $a = k \pm b$, and on $b$ in Mistral-7B, it gives FOLLOWS one token before the value is written, and in 420 fresh null worlds it never does. A Qwen2.5-1.5B control failed its accuracy gate (0.78, below 0.80; unscorable) and was replaced by Mistral (after Qwen2.5-3B failed its accuracy pilot) after that result was known, with the statistic already frozen. Three tokens before $b$ is written, and earlier, the same prompts are UNCLEAR and prefer the difference on both kinds of problem ($D_+ < 0$; $I/N$ 0.61 three tokens before in Qwen2.5-7B's development run, 0.66 in Mistral), the pattern of the Kudo positions in Section~\ref{sec:kudo}; so this pattern does not show that a value is absent.

\paragraph{An unwritten value in a pretrained model.} For a value that a pretrained model computes, uses and never writes, we used two-hop recall in Qwen2.5-7B: ``The capital of the country where the city of $X$ is located is'', after three worked examples. The country is never written. Pilots without any learned subspace set the ground truth (Appendix~\ref{app:twohop}). Replacing the whole state at the city's last token with that of a city in another country moves the answer to that country's capital in 0.82--0.84 of prompts (layers 4--12), so the country is used there. At the prompt's last token a nearest-class-mean readout gives the country at 0.92--1.00 from layer 12 on, but the same patch does nothing through layer 20. The clamp replaces the country's subspace (24 dimensions, spanned by class means of 25 countries and projected off the class means of the city name's token embeddings, the copy) after layers 8, 12 and 16. On 1,038 cities the model knows (pre-registered), it moves the answer to the other country's capital in 79\% of pairs at the city's token (USED, +0.789 [0.732, 0.841] over the random subspace; whole state 0.85; the copy's subspace 0.00), and in none at the last token (NOT-USED). There the whole state does nothing either, so this only rules out effects through dormant directions \citep{makelov2023}; at the city's token the copy's subspace does nothing (0.00), and same-country pairs keep the answer. So in a pretrained model the clamp finds an unwritten value that is used, and is silent where the same value is readable and unused. The signal is strong, though: the copy projection keeps 0.99 of the country's subspace, so this shows no power at Ouro's or Kudo's signal level.

\paragraph{An unwritten value at a distance.} For a value that is computed, used and never written, we trained 6-layer transformers from scratch (5 seeds) on ``$x \circ z$ ; [6 fillers] $w$ ='' with answer $T[x \circ z] + w$ (mod 23), $T$ a fixed permutation, and negative controls on $T_1[x] + T_2[z] \circ w$, which never need $s = x \circ z$. Distributed alignment search \citep{geiger2024finding}, run before any protocol statistic, places $s$ itself (not the operands) at ``;'', eight tokens before the answer, and at most weakly elsewhere (Appendix~\ref{app:v230}). In the pre-registered test the clamp is USED there in all five models (+0.60 to +0.98 over a random subspace; also with no fillers), and not in any negative control or where $s$ is absent. The share swap is FOLLOWS there in all five, but also at every other position, where nothing uses $s$ but a probe reads it beyond the copies (post hoc), and in one of 270 null worlds, at a tiny needed share, so by its pre-registered rule it is not specific. Here a FOLLOWS marks presence, and only the clamp tests use.

\section{Application: an unwritten intermediate in a looped model}\label{sec:ouro}

Studies 1 and 2 used two-step word problems whose answer is $a+b-c$, so the intermediate $a+b$ is never written (``Maya has 30 stickers. Her friend gives her 30 more stickers. She then uses 13 stickers on a card. How many stickers does Maya have left?''), 7,200 problems from 12 templates each. At P1 after loop 4, $a+b$ reads +0.21 (Study 1, mod 10) and +0.22 (Study 2, mod 5) above its copy ceiling, and better than its sign-flipped twin $a-b$ on every one of the 24 templates (post hoc); at P0, before the answer starts, it is not readable. Study 1 missed its pre-registered bar and Study 2's bar was set after it (Appendix~\ref{app:ouro}). Every problem needed $a+b$, so a fixed preference for sums would also explain these results. Study 3 swaps the operation.

\paragraph{Study 3: the operation swap.} Its 12 new templates come in two versions that differ only in the first step's phrase: ``gave her $B$ more stickers'' gives $a+b-c$, and ``took $B$ of her stickers'' gives $a-b-c$. All gates passed (answer-only accuracy 0.67 and 0.77). The accuracy-based swap gives FOLLOWS ($I = +0.157$, Table~\ref{tab:ouro}), but that swap later failed its positive control. Re-scored with the validated share swap (pre-registered after its validation), P1 is still FOLLOWS, with the copies of $a$, $b$, $c$ and the answer's digit regressed out: $a+b$ explains 0.006 of the remaining variance and $a-b$ 0.004 on plus problems, and 0.001 and 0.008 on minus problems, against 0.001 for random functions of the operands on average ($I/N = 0.63$, bar 0.30). P1 is the answer's tens digit, which the carry and borrow fix within a family; with them regressed too (pre-registered) it stays FOLLOWS ($I/N = 0.57$; the sum's lead on plus problems falls to 0.001), against a bar of 0.25 from null worlds rebuilt on this design down to Ouro's signal level (none of 270 passes). P0, before the answer starts, is UNCLEAR. The effect is small: under 1\% of the remaining variance, against 11--15\% where a value is about to be written (Section~\ref{sec:control}). No probe label can separate ``$a+b$ was formed and the answer computed from it'' from ``the answer was formed and $c$ added back'', since on units digits they are the same function. Nor does training order in 18 small transformers trained from scratch (Appendix~\ref{sec:v220}). Where $c$ is not yet read, at the end of the sentence that gives $b$, the share swap is UNCLEAR, though it would likely see a running total as strong as P1's signal, and even the whole state there leaves the answer unchanged (pre-registered; Appendix~\ref{sec:study3}).

\begin{table}[t]
\centering
\caption{Ouro-1.4B, Study 3, P1 (pre-registered). Upper: excess over the copy ceiling (mod 5) after loop 4, with a 97.5\% interval over twin families, the accuracy-based swap and the validated share swap ($I$ as a share of the needed label's, averaged over the middle and end of loops 3 and 4). Lower: share of 2,000 pairs in which the written digit becomes the partner's when the answer's subspace is replaced (positive control), and $a+b$'s hit rate minus the larger of $a-b$'s and a random subspace's, with its pre-registered interval (99.4\% and 99.2\%). The labels follow the pre-registered rule, whose comparison with $a-b$ proved weak (Appendix~\ref{app:v230}); against the random subspace alone both are UNCLEAR (post hoc).}
\label{tab:ouro}
\small
\setlength{\tabcolsep}{3pt}
\begin{tabular}{@{}lcc@{}}
\toprule
 & plus problems & minus problems \\
\midrule
excess of $a+b$ & +0.207 & +0.078 \\
excess of $a-b$ & +0.082 & +0.267 \\
accuracy swap $I$ & \multicolumn{2}{c}{+0.157 [0.141, 0.172]} \\
mixing null, largest $I$ & \multicolumn{2}{c}{+0.14} \\
share swap $I/N$ & \multicolumn{2}{c}{0.63 (bar 0.30): FOLLOWS} \\
\quad with carry, borrow & \multicolumn{2}{c}{0.57 (bar 0.25): FOLLOWS} \\
\midrule
 & patch, loop 3 & clamp, loops 1--3 \\
\midrule
answer subspace & 0.902 & 0.914 \\
$a+b$ minus control & $-0.015$ & $-0.016$ \\
 & [$-0.027$, $-0.004$] & [$-0.034$, +0.001] \\
label & NOT-USED & NOT-USED \\
\bottomrule
\end{tabular}
\end{table}

\paragraph{Is it used?} For two plus problems from the same family whose units digits of $a+b$ differ but share their parity, we replace $x$'s component at P1 with $x'$'s and read the digit Ouro writes. Patching the answer's subspace after loop 3 moves it to $x'$'s digit in 90\% of pairs; patching $a+b$'s does so in 1.7\%, less often even than $a-b$'s (3.2\%), so the label is NOT-USED (Table~\ref{tab:ouro}). An intact answer could override a weak $a+b$ used to form it, so a further pre-registered study clamped the component at the ends of loops 1--3, before the answer forms: the answer's clamp moves the written digit in 91\% of pairs and $a+b$'s in about 4\%, less often than $a-b$'s, also in pairs with equal answers. These interventions predate the copy projection. With its fitting data degraded to Ouro's signal level, a clamp of this form still moves the answer in 44--98\% of pairs in the small models (USED in 5 of 5; Appendix~\ref{app:v230}), but Ouro's rule, comparing with $a-b$'s subspace, is weak there. Against random alone (post hoc), the loop-3 patch and the clamp are slightly above chance ($+0.017$ [0.008, 0.027], $+0.039$ [0.027, 0.051]), UNCLEAR by the rule, so at P1 the value is readable and its use unresolved.

\section{Application: are sub-answers derived before the chain of thought?}\label{sec:kudo}

\citet{kudo2026} conclude that language models ``have not derived an answer at the moment when input was fed into the model'' and compute (sub-)answers while writing the chain of thought. Their evidence is a standard probe at each position and layer, which before the chain reads the sub-answer at 51\% and the answer at 18\%, against 95--100\% inside it and a 90\% threshold, and patching. We applied the protocol to their task (Section~\ref{sec:control}) in five pre-registered studies, in their format with three worked examples, in Qwen2.5-7B. Here $a$ is the answer and $b$ the sub-answer. Families vary $d_1$ and $d_2$ over every valid pair, in a plus and a minus version; $k$ varies across families, which separates $b$ from the answer. We read the end of $b$'s equation and the last prompt token and, without a chain, the token just before the answer digit, which we call the answer position (Qwen gives the space before a digit its own token; that space is the answer position). Without a chain, a third study uses pairs with equal answers, a fourth replaces the whole state at the five positions between $b$'s equation and the answer position, and a fifth repeats that with pairs whose answers differ.

\begin{table*}[t]
\centering
\caption{The re-test of \citet{kudo2026}, Qwen2.5-7B. Beyond copies: the needed form's excess over the copy ceiling (layers 12--24). Share swap: $I$ (layers 16--28; with no chain, the answer regressed out too); every position is UNCLEAR, with the difference preferred on both kinds of problem. Short-chain swap values (a) were seen while the swap was developed. Clamp (1,000 pairs): share of pairs in which the model writes the other problem's $b$ (short chain) or answer (no chain) after replacing $b$'s directions with the copy projected out (with no chain, the copy and the answer), the copied digit's directions, or the whole state.}
\label{tab:kudo}
\small
\begin{tabular}{@{}llccc@{}}
\toprule
position & chain & beyond copies & share swap $I$ & clamp: $b$ / copy / whole \\
\midrule
end of $b$'s equation & theirs & +0.38 & +0.020 & -- \\
 & short & +0.37 & +0.018$^a$ & 0.02 / 0.96 / 0.96 \\
last prompt token & theirs & +0.14 & +0.010 & -- \\
 & short & +0.12 & +0.007$^a$ & unscorable (whole 0.00) \\
\midrule
end of $b$'s equation & none & +0.38 & +0.006 & 0.01 / 0.83 / 0.86 \\
just before the answer & none & +0.27 & +0.015 & 0.03 / 0.24 / 0.85 \\
last prompt token & none & +0.14 & +0.013 & unscorable (whole 0.00) \\
\bottomrule
\end{tabular}
\end{table*}

\paragraph{With a chain of thought.} Their probe's result is reproduced, at a higher level: before the chain it reads $b$ at 0.64 (their 50.5\%) and $a$ at 0.32 (17.9\%). The protocol shows what that readability is (Table~\ref{tab:kudo}). The operands are combined beyond the copy ceiling at both positions (+0.12 to +0.38), but the share swap does not find it following the operation: it is UNCLEAR at both, with the difference preferred on plus and minus problems alike (on plus problems random functions of the operands explain as much). Nor does the chain use such a value. In their format the chain restates $b$'s equation, and replacing the whole state at either position from layer 12 on never changed the written $b$ in a 40-pair check (not pre-registered). In a short format without the restated equation (``$b = 5$, $a = 6$''), replacing the state at the end of $b$'s equation changes the written $b$ in 96\% of pairs, and so does replacing only the copied digit's directions (96\%), while $b$'s directions with the copy projected out do so in 2\% (+0.021 [0.011, 0.033] over the random clamp; UNCLEAR by the clamp's rule; descriptive, since no position was FOLLOWS).

\paragraph{Without a chain.} Answering directly (their Implicit setting), Qwen2.5-7B is right on 69\% of random problems like theirs (their 77.8\%; 76\% on our families). Their probe reads $b$ at 0.74 just before the answer (theirs 69.1\%). Here the answer is itself a confound, since within a family it fixes $b$, so the share swap regresses the answer out too and the clamp projects $b$'s directions off both. The operands are again combined at every position (+0.14 to +0.38), but with the answer regressed out as well, the share swap is UNCLEAR at every position, with the same preference for the difference and a small part that follows the operation. Replacing the whole state changes the answer to the partner's in 85--86\% of pairs at the two scorable positions (the last prompt token is unscorable). At the end of $b$'s equation the copied digit's directions alone do so in 83\%; just before the answer, the answer's directions do so in 46\%; $b$'s directions with the copy and the answer projected out, in 1\% (NOT-USED) and 3\%. The 3\% is above the random clamp ($+0.025$ [0.014, 0.039]; UNCLEAR, descriptive). The projections keep only 35--54\% of $b$'s directions, too few for the power the clamp has shown (Appendix~\ref{app:v230}), so these results say little. Within a family the answer fixes $b$, so this design cannot separate the two either.

\paragraph{Equal answers.} A third study removed the answer from the clamp. New families vary $k$ too, and each pair shares the answer while differing in $b$ (and $k$), so the answer's code carries no difference and a used $b$ predicts an answer neither prompt has, $x$'s $k$ with $x'$'s $b$. Only the copies need projecting out, which keeps 85\% of $b$'s directions. At the end of $b$'s equation the whole state gives that prediction in 87\% of 1,000 pairs and $d_2$'s copy alone in 88\%; $b$'s directions beyond the copies in 4\% (+0.038 [0.023, 0.056] over random; UNCLEAR). That token is $d_2$ itself, so what the answer takes from it is mostly the copy.

\paragraph{Relay tests after $b$'s equation.} A whole-state patch needs no subspace, so it has no power problem of the clamp's kind; the last two studies use it, as ordinary multi-position activation patching \citep{meng2022locating,zhang2024patching}. Both problems of a pair are answered correctly; one type differs in $d_1$, so the end of $b$'s equation is the same token in both prompts, the other in $d_2$. After a pilot (Appendix~\ref{app:kudo}), the fourth study replaces the five positions between $b$'s equation and the answer position (`` ;'', the question's letter, ``?'', and `` a'' and `` ='' of the target line `` a = A'' that the model completes) after every layer at once, in pairs that share the answer and differ in $k$. Provided the operand's own token moves the answer to the prediction in at least 30\% of pairs, the result is HELD if the relay does so in at least 10\% with the interval above zero and NOT-HELD if the upper bound is below 2\%. The relay gives it in 0.3\% and 0.2\% of 1,000 pairs per type (NOT-HELD), the operand's own token in 90\% and 88\% (Table~\ref{tab:relay}). Pairs with equal answers, though, see $b$ carried apart from $k$, not a finished answer. The fifth study, run after a draft review raised this, uses pairs that share $k$, so that a carried $b$, operand or answer all predict the partner's answer: the relay gives it in 8.1\% and 8.3\% ([0.061, 0.103] and [0.062, 0.107]; UNCLEAR, above zero and below the bar), the operand's token in 99.8\% and 99.9\%.

\paragraph{Where the 8\% sits (post hoc).} A map on the fifth study's pairs puts most of it at `` ='', the target line's token before the answer position: alone it gives 5.8\% and 5.4\%, the other four positions together 1.1\% and 2.0\%; `` ;'', the question's letter and ``?'' each give at most 1.4\% alone, and the target line's `` a'' none. Among the single layers tested (8--23), patches of the five positions matter only from layer 20 (at most 0.8\% after layers 8, 12 and 16; 2--4\% after 20, 22 and 23). Since the pairs share $k$, input copies held with $k$ there would look the same as a carried answer, so the patch shows only that `` ='' holds something, used with $k$, that gives the partner's answer. The end of $b$'s equation, outside the five, gives the predicted answer in 6\% (equal answers) and 7\% (different answers) of pairs that differ in $d_1$, as much as `` ='' alone. Each operand's token stops mattering after layer 23 of 28, and the answer position holds the partner's answer after layer 23 (76--95\%, different answers).

\begin{table}[t]
\centering
\caption{Without a chain: share of pairs (1,000 per type, both answered correctly) in which the answer becomes the one a used $b$ would give, after replacing the whole state with the partner's; upper rows with equal answers, lower rows with different answers (same $k$; hit = the partner's answer). Relay: all five positions between $b$'s equation and the token before the answer, after every layer (pre-registered; upper bound of the 97.5\% interval in brackets). Operand: the operand token where the two prompts differ (in the upper rows they also differ in $k$), after every layer (the positive control) or after one layer; other operand: the end of $b$'s equation or $d_1$'s token. Answer position: the token before the answer digit, after one layer (descriptive).}
\label{tab:relay}
\footnotesize
\setlength{\tabcolsep}{3pt}
\begin{tabular}{@{}lcc@{}}
\toprule
pairs differ in & $d_1$ & $d_2$ \\
\midrule
relay, every layer & 0.003 [0.007] & 0.002 [0.006] \\
\quad label & NOT-HELD & NOT-HELD \\
operand, every layer & 0.90 & 0.88 \\
other operand, every layer & 0.06 & 0.01 \\
operand, after layer 22 & 0.70 & 0.40 \\
operand, after layer 23 & 0.02 & 0.01 \\
answer position, after layer 22 & 0.06 & 0.26 \\
\midrule
relay, different answers & 0.081 [0.103] & 0.083 [0.107] \\
\quad label & UNCLEAR & UNCLEAR \\
other operand, different answers & 0.069 & 0.000 \\
`` ='' alone (post hoc) & 0.058 & 0.054 \\
all but `` ='' (post hoc) & 0.011 & 0.020 \\
\bottomrule
\end{tabular}
\end{table}

\paragraph{What this says about their claim.} Weaker tests point to an early sub-answer. Their probe reads $b$ at 0.64 before the chain, and without one at up to 0.58 at the last prompt token (majority class 0.20), 0.63 at the end of $b$'s equation and 0.74 at the answer position (after the last layer, where it reads the answer at 0.85); the operands are combined beyond the copy ceiling everywhere; and a clamp that keeps the copy moves the written $b$ in 30\% of pairs and the answer in 28--35\%. The share swap is UNCLEAR everywhere, as three tokens before $b$ is written in the control (Section~\ref{sec:control}), and projecting out the copy removes nearly all of the clamp's effect. Without a chain the whole-state tests say more. In correctly answered direct-answer problems they fit the conclusion of \citet{kudo2026} for the sub-answer: $b$ is carried from its equation at most weakly (4--7\%) and not on its own past it. For the answer, the different-answer relay test is UNCLEAR (8\%, below our bar), so the equal-answer caveat stays: something that depends on $b$ ($b$ with $k$, the answer, or copies of the inputs held with $k$) is carried weakly before the answer position. The post hoc map puts most of it at `` ='', a format token of the target line just before the answer position. If the target line counts as output, as in their format, this too fits their conclusion that answers are not derived when the input is read; our pre-registered reading counted `` ='' as before the answer position. Whether $b$ is formed as a separate step the protocol cannot tell, and their probe, which reads $b$ at 0.58 at the last prompt token, whose state the answer does not use, could not show which positions are used. With a chain the protocol cannot tell whether $b$ is held before the chain.

\section{Discussion}\label{sec:disc}

In each setting where we could check both, some position holds a readable value that the output does not use: the bridge country at the last token of a two-hop prompt (0.92--1.00 from layer 12, unused through layer 20), the sub-answer of \citet{kudo2026} at the last prompt token before a direct answer (up to 0.58, majority class 0.20; Section~\ref{sec:kudo}), and $s$ at the fillers of the small models (0.63--1.00; Appendix~\ref{app:v230}). In each setting the value is also used somewhere: at the city's token, at the operands' own tokens, at ``;''. A probe does not tell these positions apart \citep{ravichander2021probing,elazar2021amnesic}; an intervention does, and a null from a subspace intervention means something only next to a positive control at the same distance and signal level.

In this model a direct two-step answer comes together late (the operands' tokens stop mattering after layer 23 of 28, and their probe reads the answer at the answer position at 0.42, 0.77 and 0.85 after layers 20, 24 and 28), whereas in two-hop recall the bridge is held at the city's token from layer 4 on. This comes from one model and single-digit problems, and a late read of the operands also fits a memorised lookup over $(k, d_1, d_2)$, so we do not generalise it.

\section{Recommendations}\label{sec:recs}

We suggest seven practices: test presence with minimal pairs before reading a failed probe as absence; score a candidate against a copy ceiling that includes every input, and the answer and the carry where they fix the token read; swap the operation, and judge the swap on shares of variance against a bar set by synthetic mixtures on the study's own design; do not read an UNCLEAR swap far from where a value is written as absence; read a FOLLOWS as presence, not use; project the copy out before clamping, with a positive control at the same distance, signal level and share kept; and before reading a weak clamp as absence, replace the whole state at every position between the inputs and the answer position at once, both in pairs where a carried value predicts an output neither prompt gives and in pairs with different outputs, since the first kind cannot see a carried answer (here only the second kind showed a weak carry, 8\%, UNCLEAR).

\section{Limitations}\label{sec:lim}

The probing results rest on four models, and the computation studies on Ouro-1.4B and Qwen2.5-7B, each with one kind of two-step arithmetic read where answers begin. In Ouro, Study 2's bar was set after Study 1's near-miss, several comparisons were run afterwards on saved states, and the signal cannot be told apart from the answer with $c$ added back (Appendix~\ref{app:status}). The swap's bar rests on five kinds of synthetic mixing; real states may mix differently. The interventions test a few positions and subspaces spanned by class means: a value held elsewhere and read through attention, or in a code outside those subspaces, would be missed. In a pretrained model the clamp's power is shown for written values read at most six tokens later and for a strong unwritten bridge entity (0.99 of it kept by the projection), in Qwen2.5-7B; at Ouro's signal level only in small models trained from scratch, by degrading the clamp's fitting data, and for the form used for \citet{kudo2026} only when the projection keeps 55\% or more of the value, which its clamps do with a chain (75--85\%) and in the equal-answer study (85\%), but not in the first no-chain study (35--54\%). The direct-answer result for \citet{kudo2026} rests on whole-state patches in one model, on a restricted item set (single digits, $d_2 \le d_1$, $d_1 + d_2 \le 7$) and correctly answered problems. With different answers the relay test is UNCLEAR, so a weakly carried answer (or $b$ with $k$) before the answer position is not ruled out; the map placing it at `` ='' is post hoc and times it only for the five positions together; the relay design followed its own pilot; and the protocol cannot say whether $b$ is formed as a separate step. The share swap was validated in two pretrained models and two steps of one task, with one control replaced after the first failed its accuracy gate, and in the small models it was not specific; Ouro, where it gives the paper's one FOLLOWS, is not among them. The re-test of \citet{kudo2026} uses one of their nine models and one task level.

\section{Related work}\label{sec:related}

\textbf{Looped models and latent reasoning.} The answer-only accuracy of the looped Huginn \citep{geiping2025} stays flat across recurrence counts, and a logit lens shows no intermediate values \citep{lu2025}; an unreviewed repository reports the same for Ouro-1.4B-Thinking \citep{mildrgb2026}, and \citet{kirin2026} probes Ouro for whether it will answer correctly. Ouro's state is re-derived within each loop \citep{wangreid2026,huang2026partii}, and in models pretrained from scratch at equal depth looping brings no reasoning advantage \citep{schwethelm2026a}. In theory a looped transformer can simulate $T$ chain-of-thought steps in $T$ loops \citep{saunshi2025reasoning}, and models trained to drop written steps can compute intermediates internally \citep{deng2024explicit,hao2025training}, though implicit multi-step arithmetic in models trained from scratch relies on shortcuts \citep{lin2025}. In standard models the first hop of a two-hop question is often resolved latently, the full chain less reliably \citep{yang2024latently,biran2024hopping}, and the bridge entity is resolved at the first entity's tokens before the last token completes the second hop \citep{biran2024hopping}, as our pretrained control finds; last-token probes rarely detect intermediate steps in prompted implicit reasoning \citep{yu2024}, and sub-answers are computed while the chain is written \citep{kudo2026}.

\textbf{Probing and its limits.} We build on control tasks and selectivity \citep{hewitt2019}, conditional probing \citep{hewitt2021conditional} and difference-in-means probing \citep{marks2023}; our learning curves are close in spirit to information-theoretic and minimum-description-length probing \citep{voita2020mdl,pimentel2020information}. \citet{park2024} define concept directions from counterfactual pairs, as our minimal pairs do. Probing critiques \citep{belinkov2022probing} stress that a readable property need not be used \citep{ravichander2021probing,elazar2021amnesic}; a probe can also fail because of how its target is framed \citep{nanda2023emergent}. Interchange interventions \citep{geiger2021causal,geiger2024finding} and causal mediation with counterfactual inputs \citep{vig2020investigating,stolfo2023,hanna2023greaterthan} test whether a variable is used; subspace patches can mislead by acting through dormant directions \citep{makelov2023}, and the copy projection guards against a related risk. \citet{ye2024} read needed quantities at the question's end in models trained from scratch. To our knowledge, no earlier work reads a copied token at a summary position with minimal pairs while a standard probe fails, or tests a computed value against copies, an operation swap and calibrated mixtures.

\textbf{Number representations.} Numbers are encoded partly linearly \citep{zhuf2024} and partly by periodic features with periods such as 2, 5 and 10 \citep{zhou2024fourier,kantamneni2025trigonometry,nanda2023progress}. Arithmetic can come from a bag of heuristics \citep{nikankin2025arithmetic}, which fits the generic mixing we see; small transformers trained on addition build carry features \citep{quirke2024addition}.

\section*{Impact Statement}

This paper presents work whose goal is to advance the field of machine learning, specifically methods for reading what language models compute internally. Such methods are used to monitor models, and the paper's main point is a caution for that use: a failed probe does not show that a value is absent, and a successful one does not show that it was computed or used. We see no other societal consequences that need discussion here.

\section*{Reproducibility}

Code, pre-registrations, frozen file hashes, analysis outputs and decision rules for every experiment are in a public repository (\url{https://github.com/Aur-Shalev-Merin/present-but-hard-to-read}), including per-item results for the post-hoc comparisons. Every pre-registration was committed to a private development repository before its run or analysis; all but the two accuracy checks of Appendix~\ref{app:acc} were also pushed before it, though for five studies the first-run time could not be retrieved. The public repository lists each freeze commit, when it reached GitHub, and the first run where recorded. Each experiment's document records its predictions, its results, and every change made before data.

\bibliography{main}

\begin{thebibliography}{52}
\providecommand{\natexlab}[1]{#1}
\providecommand{\url}[1]{\texttt{#1}}
\expandafter\ifx\csname urlstyle\endcsname\relax
  \providecommand{\doi}[1]{doi: #1}\else
  \providecommand{\doi}{doi: \begingroup \urlstyle{rm}\Url}\fi

\bibitem[Belinkov(2022)]{belinkov2022probing}
Belinkov, Y.
\newblock Probing classifiers: Promises, shortcomings, and advances.
\newblock \emph{Computational Linguistics}, 48\penalty0 (1):\penalty0 207--219,
  2022.
\newblock \doi{10.1162/coli_a_00422}.

\bibitem[Biran et~al.(2024)Biran, Gottesman, Yang, Geva, and
  Globerson]{biran2024hopping}
Biran, E., Gottesman, D., Yang, S., Geva, M., and Globerson, A.
\newblock Hopping too late: Exploring the limitations of large language models
  on multi-hop queries.
\newblock In \emph{Proceedings of the 2024 Conference on Empirical Methods in
  Natural Language Processing}, pp.\  14113--14130. Association for
  Computational Linguistics, 2024.
\newblock \doi{10.18653/v1/2024.emnlp-main.781}.

\bibitem[Cobbe et~al.(2021)Cobbe, Kosaraju, Bavarian, Chen, Jun, Kaiser,
  et~al.]{cobbe2021}
Cobbe, K., Kosaraju, V., Bavarian, M., Chen, M., Jun, H., Kaiser, L., et~al.
\newblock Training verifiers to solve math word problems.
\newblock arXiv preprint arXiv:2110.14168, 2021.

\bibitem[Deng et~al.(2024)Deng, Choi, and Shieber]{deng2024explicit}
Deng, Y., Choi, Y., and Shieber, S.
\newblock From explicit {CoT} to implicit {CoT}: Learning to internalize {CoT}
  step by step.
\newblock arXiv preprint arXiv:2405.14838, 2024.

\bibitem[Elazar et~al.(2021)Elazar, Ravfogel, Jacovi, and
  Goldberg]{elazar2021amnesic}
Elazar, Y., Ravfogel, S., Jacovi, A., and Goldberg, Y.
\newblock Amnesic probing: Behavioral explanation with amnesic counterfactuals.
\newblock \emph{Transactions of the Association for Computational Linguistics},
  9:\penalty0 160--175, 2021.
\newblock \doi{10.1162/tacl_a_00359}.

\bibitem[Elhage et~al.(2022)Elhage, Hume, Olsson, Schiefer, Henighan, Kravec,
  Hatfield-Dodds, Lasenby, Drain, Chen, Grosse, McCandlish, Kaplan, Amodei,
  Wattenberg, and Olah]{elhage2022}
Elhage, N., Hume, T., Olsson, C., Schiefer, N., Henighan, T., Kravec, S.,
  Hatfield-Dodds, Z., Lasenby, R., Drain, D., Chen, C., Grosse, R., McCandlish,
  S., Kaplan, J., Amodei, D., Wattenberg, M., and Olah, C.
\newblock Toy models of superposition.
\newblock arXiv preprint arXiv:2209.10652, 2022.

\bibitem[Feng \& Steinhardt(2024)Feng and Steinhardt]{feng2024binding}
Feng, J. and Steinhardt, J.
\newblock How do {Language} {Models} {Bind} {Entities} in {Context}?
\newblock In \emph{The Twelfth International Conference on Learning
  Representations ({ICLR})}, 2024.

\bibitem[Feucht et~al.(2024)Feucht, Atkinson, Wallace, and Bau]{feucht2024}
Feucht, S., Atkinson, D., Wallace, B., and Bau, D.
\newblock Token erasure as a footprint of implicit vocabulary items in {LLMs}.
\newblock arXiv preprint arXiv:2406.20086, 2024.

\bibitem[Geiger et~al.(2021)Geiger, Lu, Icard, and Potts]{geiger2021causal}
Geiger, A., Lu, H., Icard, T., and Potts, C.
\newblock Causal abstractions of neural networks.
\newblock In \emph{Advances in Neural Information Processing Systems},
  volume~34, pp.\  9574--9586, 2021.

\bibitem[Geiger et~al.(2024)Geiger, Wu, Potts, Icard, and
  Goodman]{geiger2024finding}
Geiger, A., Wu, Z., Potts, C., Icard, T., and Goodman, N.
\newblock Finding alignments between interpretable causal variables and
  distributed neural representations.
\newblock In \emph{Proceedings of the Third Conference on Causal Learning and
  Reasoning}, volume 236 of \emph{Proceedings of Machine Learning Research},
  pp.\  160--187. PMLR, 2024.

\bibitem[Geiping et~al.(2025)Geiping, McLeish, Jain, Kirchenbauer, Singh,
  Bartoldson, Kailkhura, Bhatele, and Goldstein]{geiping2025}
Geiping, J., McLeish, S., Jain, N., Kirchenbauer, J., Singh, S., Bartoldson,
  B.~R., Kailkhura, B., Bhatele, A., and Goldstein, T.
\newblock Scaling up test-time compute with latent reasoning: A recurrent depth
  approach.
\newblock In \emph{Advances in Neural Information Processing Systems},
  volume~38, 2025.

\bibitem[Hanna et~al.(2023)Hanna, Liu, and Variengien]{hanna2023greaterthan}
Hanna, M., Liu, O., and Variengien, A.
\newblock How does {GPT-2} compute greater-than?: {Interpreting} mathematical
  abilities in a pre-trained language model.
\newblock In \emph{Advances in Neural Information Processing Systems},
  volume~36, pp.\  76033--76060, 2023.
\newblock \doi{10.52202/075280-3322}.

\bibitem[Hao et~al.(2025)Hao, Sukhbaatar, Su, Li, Hu, Weston, and
  Tian]{hao2025training}
Hao, S., Sukhbaatar, S., Su, D., Li, X., Hu, Z., Weston, J., and Tian, Y.
\newblock Training large language models to reason in a continuous latent
  space.
\newblock In \emph{Conference on Language Modeling (COLM)}, 2025.

\bibitem[Hewitt \& Liang(2019)Hewitt and Liang]{hewitt2019}
Hewitt, J. and Liang, P.
\newblock Designing and interpreting probes with control tasks.
\newblock In \emph{Proceedings of the 2019 Conference on Empirical Methods in
  Natural Language Processing (EMNLP)}, 2019.

\bibitem[Hewitt et~al.(2021)Hewitt, Ethayarajh, Liang, and
  Manning]{hewitt2021conditional}
Hewitt, J., Ethayarajh, K., Liang, P., and Manning, C.~D.
\newblock Conditional probing: Measuring usable information beyond a baseline.
\newblock In \emph{Proceedings of the 2021 Conference on Empirical Methods in
  Natural Language Processing}, 2021.

\bibitem[Huang et~al.(2026)Huang, Shi, Chen, Wen, Liu, Xing, and
  Ma]{huang2026partii}
Huang, B., Shi, C., Chen, J., Wen, S., Liu, Z., Xing, E., and Ma, X.
\newblock Towards looped models done right, part {II}: Rethinking at fixed
  points.
\newblock arXiv preprint arXiv:2610.06833, 2026.

\bibitem[Kantamneni \& Tegmark(2025)Kantamneni and
  Tegmark]{kantamneni2025trigonometry}
Kantamneni, S. and Tegmark, M.
\newblock Language models use trigonometry to do addition.
\newblock arXiv preprint arXiv:2502.00873, 2025.

\bibitem[Kirin(2026)]{kirin2026}
Kirin, J.
\newblock Operational proto-introspection in looped language models:
  Process-quality taps, executable branching, and the readout-control boundary.
\newblock arXiv preprint arXiv:2607.18553, 2026.

\bibitem[Kudo et~al.(2026)Kudo, Aoki, Kuribayashi, Sone, Taniguchi, Brassard,
  Sakaguchi, and Inui]{kudo2026}
Kudo, K., Aoki, Y., Kuribayashi, T., Sone, S., Taniguchi, M., Brassard, A.,
  Sakaguchi, K., and Inui, K.
\newblock {LLMs} faithfully and iteratively compute answers during {CoT}: A
  systematic analysis with multi-step arithmetics.
\newblock In \emph{Findings of the Association for Computational Linguistics:
  EACL 2026}, pp.\  1114--1153, 2026.
\newblock \doi{10.18653/v1/2026.findings-eacl.59}.

\bibitem[Levy \& Geva(2025)Levy and Geva]{levy2025}
Levy, A.~A. and Geva, M.
\newblock Language models encode numbers using digit representations in base
  10.
\newblock In \emph{Proceedings of the 2025 Conference of the Nations of the
  Americas Chapter of the Association for Computational Linguistics (NAACL)},
  2025.

\bibitem[Lin et~al.(2025)Lin, Xie, Yuan, and Yang]{lin2025}
Lin, T., Xie, J., Yuan, S., and Yang, D.
\newblock Implicit reasoning in transformers is reasoning through shortcuts.
\newblock In \emph{Findings of the Association for Computational Linguistics:
  ACL 2025}, 2025.

\bibitem[Lu et~al.(2025)Lu, Yang, Lee, Li, and Liu]{lu2025}
Lu, W., Yang, Y., Lee, K., Li, Y., and Liu, E.
\newblock Latent chain-of-thought? {D}ecoding the depth-recurrent transformer.
\newblock In \emph{Workshop on the Application of LLM Explainability to
  Reasoning and Planning, COLM 2025}, 2025.

\bibitem[Makelov et~al.(2024)Makelov, Lange, and Nanda]{makelov2023}
Makelov, A., Lange, G., and Nanda, N.
\newblock Is this the subspace you are looking for? {An} interpretability
  illusion for subspace activation patching.
\newblock In \emph{International Conference on Learning Representations
  (ICLR)}, 2024.

\bibitem[Marks \& Tegmark(2023)Marks and Tegmark]{marks2023}
Marks, S. and Tegmark, M.
\newblock The geometry of truth: Emergent linear structure in large language
  model representations of true/false datasets.
\newblock arXiv preprint arXiv:2310.06824, 2023.

\bibitem[Meng et~al.(2022)Meng, Bau, Andonian, and Belinkov]{meng2022locating}
Meng, K., Bau, D., Andonian, A., and Belinkov, Y.
\newblock Locating and editing factual associations in {GPT}.
\newblock In \emph{Advances in Neural Information Processing Systems
  (NeurIPS)}, 2022.

\bibitem[{mild-rgb}(2026)]{mildrgb2026}
{mild-rgb}.
\newblock ouro-experiments.
\newblock GitHub repository, \url{https://github.com/mild-rgb/ouro-experiments}
  (not peer reviewed), 2026.

\bibitem[Mirzadeh et~al.(2024)Mirzadeh, Alizadeh, Shahrokhi, Tuzel, Bengio, and
  Farajtabar]{mirzadeh2024}
Mirzadeh, I., Alizadeh, K., Shahrokhi, H., Tuzel, O., Bengio, S., and
  Farajtabar, M.
\newblock {GSM-Symbolic}: Understanding the limitations of mathematical
  reasoning in large language models.
\newblock arXiv preprint arXiv:2410.05229, 2024.

\bibitem[Mitra et~al.(2024)Mitra, Khanpour, Rosset, and
  Awadallah]{mitra2024orcamath}
Mitra, A., Khanpour, H., Rosset, C., and Awadallah, A.
\newblock Orca-math: Unlocking the potential of {SLMs} in grade school math.
\newblock arXiv:2402.14830, 2024.

\bibitem[Moreno-Bote et~al.(2014)Moreno-Bote, Beck, Kanitscheider, Pitkow,
  Latham, and Pouget]{morenobote2014}
Moreno-Bote, R., Beck, J., Kanitscheider, I., Pitkow, X., Latham, P., and
  Pouget, A.
\newblock Information-limiting correlations.
\newblock \emph{Nature Neuroscience}, 17\penalty0 (10):\penalty0 1410--1417,
  2014.

\bibitem[Nanda et~al.(2023{\natexlab{a}})Nanda, Chan, Lieberum, Smith, and
  Steinhardt]{nanda2023progress}
Nanda, N., Chan, L., Lieberum, T., Smith, J., and Steinhardt, J.
\newblock Progress measures for grokking via mechanistic interpretability.
\newblock In \emph{International Conference on Learning Representations
  (ICLR)}, 2023{\natexlab{a}}.

\bibitem[Nanda et~al.(2023{\natexlab{b}})Nanda, Lee, and
  Wattenberg]{nanda2023emergent}
Nanda, N., Lee, A., and Wattenberg, M.
\newblock Emergent linear representations in world models of self-supervised
  sequence models.
\newblock In \emph{Proceedings of the 6th BlackboxNLP Workshop: Analyzing and
  Interpreting Neural Networks for NLP}, pp.\  16--30. Association for
  Computational Linguistics, 2023{\natexlab{b}}.
\newblock \doi{10.18653/v1/2023.blackboxnlp-1.2}.

\bibitem[Nikankin et~al.(2025)Nikankin, Reusch, Mueller, and
  Belinkov]{nikankin2025arithmetic}
Nikankin, Y., Reusch, A., Mueller, A., and Belinkov, Y.
\newblock Arithmetic without algorithms: Language models solve math with a bag
  of heuristics.
\newblock In \emph{International Conference on Learning Representations
  (ICLR)}, 2025.

\bibitem[Park et~al.(2024)Park, Choe, and Veitch]{park2024}
Park, K., Choe, Y.~J., and Veitch, V.
\newblock The linear representation hypothesis and the geometry of large
  language models.
\newblock In \emph{Proceedings of the 41st International Conference on Machine
  Learning (ICML)}, 2024.

\bibitem[Pimentel et~al.(2020)Pimentel, Valvoda, Hall~Maudslay, Zmigrod,
  Williams, and Cotterell]{pimentel2020information}
Pimentel, T., Valvoda, J., Hall~Maudslay, R., Zmigrod, R., Williams, A., and
  Cotterell, R.
\newblock Information-theoretic probing for linguistic structure.
\newblock In \emph{Proceedings of the 58th Annual Meeting of the Association
  for Computational Linguistics (ACL)}, 2020.

\bibitem[Quirke \& Barez(2024)Quirke and Barez]{quirke2024addition}
Quirke, P. and Barez, F.
\newblock Understanding {Addition} in {Transformers}.
\newblock In \emph{The Twelfth International Conference on Learning
  Representations ({ICLR})}, 2024.

\bibitem[{Qwen Team}(2024)]{qwen2025}
{Qwen Team}.
\newblock Qwen2.5 technical report.
\newblock arXiv preprint arXiv:2412.15115, 2024.

\bibitem[Ravichander et~al.(2021)Ravichander, Belinkov, and
  Hovy]{ravichander2021probing}
Ravichander, A., Belinkov, Y., and Hovy, E.
\newblock Probing the probing paradigm: Does probing accuracy entail task
  relevance?
\newblock In \emph{Proceedings of the 16th Conference of the European Chapter
  of the Association for Computational Linguistics: Main Volume}, pp.\
  3363--3377. Association for Computational Linguistics, 2021.
\newblock \doi{10.18653/v1/2021.eacl-main.295}.

\bibitem[Saunshi et~al.(2025)Saunshi, Dikkala, Li, Kumar, and
  Reddi]{saunshi2025reasoning}
Saunshi, N., Dikkala, N., Li, Z., Kumar, S., and Reddi, S.~J.
\newblock Reasoning with latent thoughts: On the power of looped transformers.
\newblock In \emph{International Conference on Learning Representations
  (ICLR)}, 2025.

\bibitem[Schwethelm et~al.(2026{\natexlab{a}})Schwethelm, R{\"u}ckert, and
  Kaissis]{schwethelm2026a}
Schwethelm, K., R{\"u}ckert, D., and Kaissis, G.
\newblock How much is one recurrence worth? {I}so-depth scaling laws for looped
  language models.
\newblock arXiv preprint arXiv:2604.21106, 2026{\natexlab{a}}.

\bibitem[Schwethelm et~al.(2026{\natexlab{b}})Schwethelm, R{\"u}ckert, and
  Kaissis]{schwethelm2026b}
Schwethelm, K., R{\"u}ckert, D., and Kaissis, G.
\newblock Depth-adaptive inference of looped language models via continuous
  depth batching.
\newblock arXiv preprint arXiv:2608.09444, 2026{\natexlab{b}}.

\bibitem[Stolfo et~al.(2023)Stolfo, Belinkov, and Sachan]{stolfo2023}
Stolfo, A., Belinkov, Y., and Sachan, M.
\newblock A mechanistic interpretation of arithmetic reasoning in language
  models using causal mediation analysis.
\newblock In \emph{Proceedings of the 2023 Conference on Empirical Methods in
  Natural Language Processing (EMNLP)}, 2023.

\bibitem[Vig et~al.(2020)Vig, Gehrmann, Belinkov, Qian, Nevo, Singer, and
  Shieber]{vig2020investigating}
Vig, J., Gehrmann, S., Belinkov, Y., Qian, S., Nevo, D., Singer, Y., and
  Shieber, S.
\newblock Investigating gender bias in language models using causal mediation
  analysis.
\newblock In \emph{Advances in Neural Information Processing Systems},
  volume~33, pp.\  12388--12401, 2020.

\bibitem[Voita \& Titov(2020)Voita and Titov]{voita2020mdl}
Voita, E. and Titov, I.
\newblock Information-theoretic probing with minimum description length.
\newblock In \emph{Proceedings of the 2020 Conference on Empirical Methods in
  Natural Language Processing (EMNLP)}, 2020.

\bibitem[Wang \& Reid(2026)Wang and Reid]{wangreid2026}
Wang, W. and Reid, F.
\newblock Looped transformers under the {J}acobian lens: Does the global
  workspace survive recurrence?
\newblock arXiv preprint arXiv:2609.01924, 2026.

\bibitem[Yang et~al.(2024)Yang, Gribovskaya, Kassner, Geva, and
  Riedel]{yang2024latently}
Yang, S., Gribovskaya, E., Kassner, N., Geva, M., and Riedel, S.
\newblock Do large language models latently perform multi-hop reasoning?
\newblock In \emph{Proceedings of the 62nd Annual Meeting of the Association
  for Computational Linguistics (Volume 1: Long Papers)}, pp.\  10210--10229.
  Association for Computational Linguistics, 2024.
\newblock \doi{10.18653/v1/2024.acl-long.550}.

\bibitem[Yang et~al.(2026)Yang, Han, Zhang, Wei, Shao, Guo, and
  Li]{yang2026stars}
Yang, X.-W., Han, Z., Zhang, X.-H., Wei, W.-D., Shao, J.-J., Guo, L.-Z., and
  Li, Y.-F.
\newblock Stabilizing recurrent dynamics for test-time scalable latent
  reasoning in looped language models.
\newblock In \emph{Proceedings of the 43rd International Conference on Machine
  Learning (ICML)}, volume 306 of \emph{Proceedings of Machine Learning
  Research}. PMLR, 2026.

\bibitem[Ye et~al.(2025)Ye, Xu, Li, and Allen-Zhu]{ye2024}
Ye, T., Xu, Z., Li, Y., and Allen-Zhu, Z.
\newblock Physics of language models: Part 2.1, grade-school math and the
  hidden reasoning process.
\newblock In \emph{International Conference on Learning Representations
  (ICLR)}, 2025.

\bibitem[Yu(2024)]{yu2024}
Yu, Y.
\newblock {LLMs} do not think step-by-step in implicit reasoning.
\newblock arXiv preprint arXiv:2411.15862, 2024.

\bibitem[Zhang \& Nanda(2024)Zhang and Nanda]{zhang2024patching}
Zhang, F. and Nanda, N.
\newblock Towards best practices of activation patching in language models:
  Metrics and methods.
\newblock In \emph{International Conference on Learning Representations
  (ICLR)}, 2024.

\bibitem[Zhou et~al.(2024)Zhou, Fu, Sharan, and Jia]{zhou2024fourier}
Zhou, T., Fu, D., Sharan, V., and Jia, R.
\newblock Pre-trained large language models use {F}ourier features to compute
  addition.
\newblock In \emph{Advances in Neural Information Processing Systems},
  volume~37, pp.\  25120--25151, 2024.
\newblock \doi{10.52202/079017-0792}.

\bibitem[Zhu et~al.(2024)Zhu, Dai, and Sui]{zhuf2024}
Zhu, F., Dai, D., and Sui, Z.
\newblock Language models encode the value of numbers linearly.
\newblock arXiv preprint arXiv:2401.03735, 2024.

\bibitem[Zhu et~al.(2025)Zhu, Wang, Hua, Zhang, Li, Que, Wei, Wen, Yin, Xing,
  Li, Shi, Ma, Li, Kergan, Smith, Qu, Hui, Wu, Min, Huang, Zhou, Ye, Liu, Yang,
  Shi, Lin, Zhao, Cai, Zhang, Huang, Bengio, and Eshraghian]{zhu2025ouro}
Zhu, R.-J., Wang, Z., Hua, K., Zhang, T., Li, Z., Que, H., Wei, B., Wen, Z.,
  Yin, F., Xing, H., Li, L., Shi, J., Ma, K., Li, S., Kergan, T., Smith, A.,
  Qu, X., Hui, M., Wu, B., Min, Q., Huang, H., Zhou, X., Ye, W., Liu, J., Yang,
  J., Shi, Y., Lin, C., Zhao, E., Cai, T., Zhang, G., Huang, W., Bengio, Y.,
  and Eshraghian, J.
\newblock Scaling latent reasoning via looped language models.
\newblock arXiv preprint arXiv:2510.25741, 2025.

\end{thebibliography}
\bibliographystyle{icml2026}

\newpage
\appendix
\onecolumn

\section{Answer-only accuracy by loop}\label{app:acc}

\begin{figure}[h]
\centering
\includegraphics[width=0.5\linewidth]{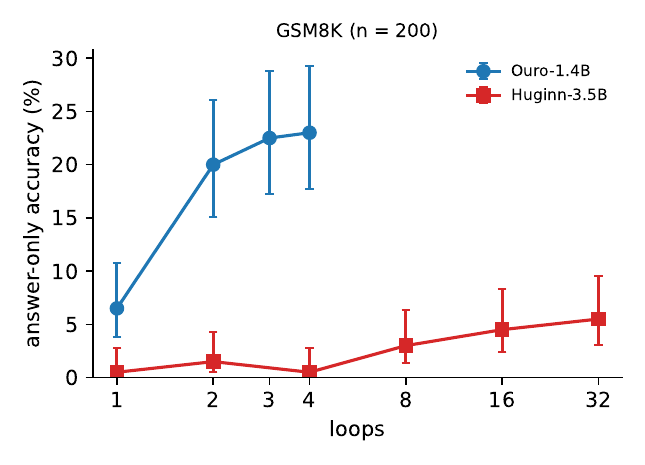}
\caption{Answer-only GSM8K accuracy by loop, with 95\% Wilson intervals.}
\label{fig:acc}
\end{figure}

On the first 200 GSM8K test problems (Figure~\ref{fig:acc}), Ouro answers 6.5\%, 20.0\%, 22.5\% and 23.0\% after loops 1 to 4. The gain over four loops is +16.5 points, 95\% CI [10.5, 22.5] by paired bootstrap; from loop 2 to loop 4 it is +3.0 [$-1.5$, +7.5]. Huginn-3.5B reaches 5.5\% at 32 recurrences, inside the interval around \citeauthor{lu2025}'s 4.9\%, which checks our harness against a published number. We do not require a newline after the answer, because Huginn's tokenizer merges line breaks; an earlier scorer that required one gave Huginn 0\%.

The loop-1-to-2 step is not specific to arithmetic. The Ouro paper's own base-model table shows the same shape on every benchmark it reports, including HellaSwag (55.2, 71.2, 74.3 at loops 1, 2, 4) and MMLU (41.2, 60.4, 67.5) \citep[Table~10]{zhu2025ouro}. Ouro's training loss covers the output of every loop \citep[Eq.~4]{zhu2025ouro}, and \citet{schwethelm2026b} exclude exits after a single loop as consistent with Ouro's exit behaviour, so part of the step may reflect a weak first exit.

On GSM-Symbolic \citep{mirzadeh2024}, where names and numbers of 100 GSM8K problems are resampled, accuracy on 500 variants rises from 2.8\% to 9.2\% (+6.4 points [3.4, 10.0], most of it at loop 2), against 10.0\% to 18.0\% on their 100 originals. A larger set of 4,200 variants from 84 templates has the same shape at a lower level (3.0\% to 7.0\%). Recall of the published problems is one explanation of the gap, but resampling also changes the numbers' size and difficulty, so the gap is at most a memorization effect. The pre-registered bar for the variant gain was half the GSM8K gain (8.25 points); it fell short and was labelled UNCLEAR. For scale, with generated reasoning Ouro scores about 43\% after one loop and 75--77\% after two to four \citep[Fig.~1]{yang2026stars}, and 78.9\% after four in the Ouro paper.

\section{The copied digit in more detail}\label{app:presence}

\subsection{Minimal pairs in Ouro}\label{sec:mp}

A standard probe at the last prompt token does not read a digit copied from the question. In a pre-registered study of intermediate values, its sanity check, the last digit of the largest number in the question, scored 40.8\%, exactly the rate of always guessing the most common digit. By position, the probe reads a copied digit perfectly on its own token, about half the time one token later (early in the first loop only), and at chance from three tokens on, consistent with the rapid loss of earlier-token information found at the last token of multi-token words in other models \citep{feucht2024}. In bare questions such as ``What is 247 plus 580?'' the same probe reads both operands at the last token at 0.96--1.00, but those prompts differ only in their numbers, which makes them close to minimal pairs.

We took one question per template and one of its multi-digit numbers, and made ten copies that differ only in that number's last digit (0--9): 137 such families from 77 distinct questions, one per template (a template's two families vary different numbers of the same question). The copies are identical token for token except one, so any difference in a later state is caused by that digit. Counterfactual minimal pairs are a standard interpretability control \citep{stolfo2023,vig2020investigating}; subtracting a family's mean removes what the family shares, the contrast that difference-in-means probing uses to find a direction \citep{marks2023}. We then repeated the test with an added sentence that cannot matter (``Unrelated fact: a nearby park has 43 benches.''), varying either its number or a colour word (``43 red benches''). Table~\ref{tab:mp} and Figure~\ref{fig:readout} give the results in Ouro.

\begin{table}[t]
\centering
\caption{Ouro, the last prompt token (chance 0.10); ranges over mid 1 and the ends of loops 1--4. The trace is the spread of a family's ten variants as a share of the spread between questions.}
\label{tab:mp}
\small
\begin{tabular}{@{}L{0.36\linewidth}ccc@{}}
\toprule
varied token & minimal-pair probe & standard probe & trace \\
\midrule
a number the problem needs & 0.92--0.97 & 0.15--0.23 & 5--7\% \\
a number in an irrelevant sentence & 0.96--1.00 & 0.14--0.30 & 3--6\% \\
a colour word in that sentence & 0.99--1.00 & 0.19--0.40 & 4--6\% \\
\bottomrule
\end{tabular}
\end{table}

Every varied token reaches the position where the answer begins, and the minimal-pair probe reads it on templates it never saw. Any changed input leaves some change in later states; what matters is that this one is in a code shared across questions, which a probe trained on other templates reads. The code is shared: a probe trained on needed numbers reads irrelevant ones at 0.78--0.97. The readout needs little of the family: applied to each test prompt minus a single other variant of its question, the same probe reads the needed digit at 0.79--0.84 (two variants, 0.87--0.91; post hoc). The irrelevant number almost never changes the first token of Ouro's answer (at most 2.6\% of families), so the trace is the copied token, not the answer in progress; this fits GSM-NoOp, where an irrelevant clause lowers accuracy \citep{mirzadeh2024}, although we did not test whether the trace causes errors. The code is not a magnitude direction \citep{zhuf2024}, consistent with per-digit codes \citep{levy2025}: a linear value component exists ($R^2$ 0.57), but the probe's errors are no closer to the true digit than random guesses (mean distance 3.8 against 3.7), and the top one to four principal components read the digit at only 0.21--0.45, against 0.93--0.97 with 256. It behaves like a code for which token was there, as the colour result suggests. A shared code is what copying predicts, since an attention head that moves a token writes roughly the same vector for it whatever the context; binding vectors that transfer across tasks have been found the same way \citep{feng2024binding}. The minimal-pair probe needs all ten variants of a question, so it is a test for presence, not a decoder. One sibling recovers much of the readout but not all of it: in a pre-registered follow-up, subtracting a single variant with a fixed other digit instead of the family mean reads 0.65--0.90, against 0.79--0.96 with all ten, in all four models of this paper; the gap is largest where the signal is weakest, and it depends on which digit serves as the reference.

\subsection{The Qwen replication, and many more questions}\label{sec:many}\label{sec:qwen}

\begin{figure}[t]
\centering
\includegraphics[width=0.6\linewidth]{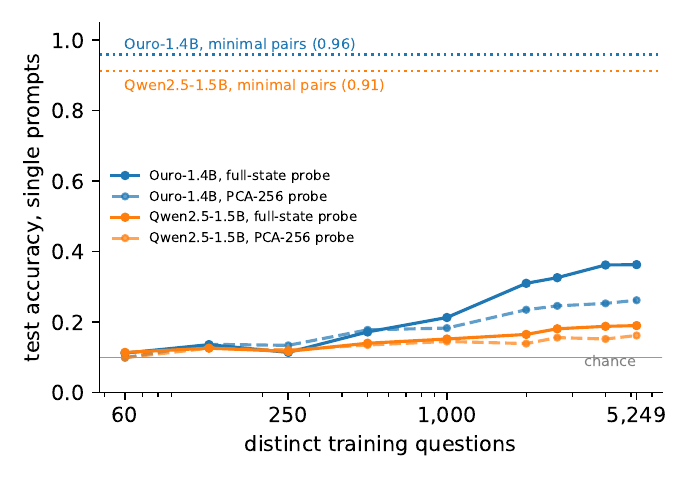}
\caption{Single-prompt probes trained on more and more distinct questions (last prompt token; best over Qwen2.5-1.5B's layers 7--28 or Ouro-1.4B's loop ends; 1,000 fixed test questions). Full-state: logistic regression on all dimensions; PCA-256: the standard probe. Dotted: minimal pairs on 150 (Qwen) and 100 (Ouro) of the test questions. The best state is picked on the test set, which favours the single-prompt probe.}
\label{fig:curve}
\end{figure}

A pre-registered study repeated these measurements in Qwen2.5-1.5B with the same families, prompt, probes and code. Its rule called the result a replication if, at the last prompt token over layers 7--28, the minimal-pair probe averaged at least 0.80, while the standard probe stayed at or below 0.30, a structured decoder at or below 0.40, and the gain from half to all training templates at or below 0.05. It is a replication (Table~\ref{tab:qwen}). The structured decoder is the best linear decoder we could build for single prompts: a mass-mean probe \citep{marks2023} built from the minimal-pair digit means and a shrunk estimate of the covariance of training prompts around them. It reads no better than the standard probe, and neither does a small MLP, nor, in Ouro, the probe without its PCA step. If the context were Gaussian with the covariance estimated from the training templates, the same decoder would read 0.53--0.92; the gap means that new templates move the state along directions the training templates did not. The claim is about decoders trained without the test question: if the covariance also includes the test question's other variants, a single prompt reads at about 0.9, which amounts to subtracting the family mean.

\begin{table}[t]
\centering
\caption{The last prompt token in a looped and a standard model, 77 distinct questions (chance 0.10). Ouro: end-of-loop states, and mid 1 where Appendix~\ref{app:state} gives it; Qwen: after layers 7, 14, 21 and 28, the pre-registered range. The gain is the standard probe's from half to all training templates.}
\label{tab:qwen}
\small
\begin{tabular}{@{}lcc@{}}
\toprule
 & Ouro-1.4B (looped) & Qwen2.5-1.5B (standard) \\
\midrule
minimal-pair probe & 0.92--0.97 & 0.75--0.98 (mean 0.88) \\
standard probe & 0.15--0.23 & 0.12--0.14 \\
structured decoder & 0.17--0.21 & 0.12--0.14 \\
MLP probe & 0.11--0.17 & 0.10--0.11 \\
gain, half to all templates & +0.015 & +0.004 \\
Gaussian ceiling of the decoder & 0.75--0.92 & 0.53--0.73 \\
\bottomrule
\end{tabular}
\end{table}

These data hold only 77 distinct questions, which leaves room for a probe that simply needs more of them. In two further pre-registered studies, one per model, we gave the model one prompt for each of 6,249 GSM8K training questions, with the units digit of the last number of two or more digits set at random, held out a fixed 1,000, and trained probes on nested subsets of the rest (Figure~\ref{fig:curve}). Minimal pairs on 150 (Qwen) and 100 (Ouro) of the held-out questions, with folds by question, read the digit at 0.91 and 0.96. Single prompts stay far below. At 5,249 training questions, 7--39 times the number of components that hold 95\% of the states' variance, the best probe, a logistic regression on the full state with its regularization inside the grid, reads 0.19 in Qwen and 0.36 in Ouro; a small MLP reads 0.11--0.14 and 0.16--0.24. Both curves are still rising. By the pre-registered rule Qwen's is partial (+0.044 over the last doubling) and Ouro's climbs (+0.060, above the rule's 0.05, though the level stays below its 0.50). The minimal pairs do not solve the single-prompt task more cheaply: they see all ten variants of the test question, so they show that the digit is present, not what it is in a new prompt. What the gap shows is that presence and single-prompt readability come apart, by a wide margin even after thousands of training questions. Why Ouro's probe climbs faster than Qwen's is open.

A further pre-registered study repeated these measurements in two larger models, Qwen2.5-7B and Ouro-2.6B (Table~\ref{tab:presence}), and extended the many-question curve to 45,249 training questions: the 5,249 GSM8K questions, then 40,000 from Orca-Math \citep{mitra2024orcamath}, after removing any that reword a test question (content-word overlap of 0.3 or more), with 1,000 Orca-Math test questions added to the GSM8K ones. Qwen2.5-7B replicates the 1.5B result by the same rule. Ouro-2.6B shows the same gap, but its minimal-pair mean falls just short of the rule's 0.80, because the digit fades after the first loop (0.92 there, 0.72--0.78 after), so by the rule its result is mixed. With 45,249 questions the single-prompt probes stay well below minimal pairs on the same test questions: 0.31 in Ouro-2.6B, 0.23--0.29 at Qwen2.5-7B's layers 7--21, and 0.52 at its last layer, the one state where a single prompt reads the digit about half the time; that layer already reads 0.43 with the GSM8K questions alone. Averaged over layers and probes, the last doubling adds +0.008 [$-0.001$, 0.018] in Qwen and +0.004 [$-0.006$, 0.014] in Ouro on the GSM8K test questions, and still +0.014 and +0.018 on the Orca-Math ones. By the pre-registered rules Qwen's curve is partial and Ouro's is shift-limited: the added questions stopped helping on GSM8K but still help a little on their own distribution, so the result does not say that more data cannot help. Removing the test questions that have a reworded twin among the training questions changes the best probe by at most 0.01.

\subsection{A small change, not a crowded one}\label{sec:swamp}

A trace at a few percent of the variance would still be easy to read if questions hardly varied along its directions. They vary a lot. Take the digit's directions to be the 9-dimensional span of the family-centred digit means, and its \emph{overlap} to be the context's variance per dimension inside that span over its average per dimension. The overlap is 6.5--11.7 in Ouro and 15--24 in Qwen, so within its own subspace the digit's signal is at most a few percent of the variance around it (Ouro by state in Appendix~\ref{app:state}). The overlap is not specific to the digit. A subspace fitted with the digit labels shuffled within each family, which carries no shared code, has about the same (ratio 0.83 in Ouro and 1.01 in Qwen), and a colour word and a number varied in the added sentence are each about as crowded as their own shuffled-label baselines, or less. Any change within a question lands on the directions along which questions vary most.

What differs between changes is their size (Table~\ref{tab:perturb}). In Qwen we also replaced the digit with one of ten random words, or added one of ten random vectors of a typical token embedding's size to its embedding. All three changes land on equally busy directions: by the pre-registered measure, the word swap's overlap is 0.89 times the digit's and the noise's 1.07. Their signal-to-noise inside their own subspaces differs by orders of magnitude, and readability follows it: a standard probe reads the word swap at 0.24--0.74, and the digit and the noise at 0.11--0.15. The input embeddings show where part of the difference starts. In both models the ten digits sit about half as far apart as random words do (0.70 against 1.37 for 60 random words in Qwen, 0.88 against 1.54 in Ouro; the ten swap words of Table~\ref{tab:perturb} are 1.42 apart). But the start does not explain all of it: the word and noise variants start equally far apart, and only the word change reaches the last token, while the noise arrives as small as the digit or smaller. These comparisons are descriptive. A pre-registered follow-up measured the linear response of the last token to small changes at the digit's input embedding (Appendix~\ref{app:state}). To first order, the last token is about as sensitive to random input directions as to directions in which token embeddings differ (gain ratio 1.2--1.5), and the word swap's effect is almost entirely nonlinear: its linear prediction is orthogonal to the actual change and 4.5--15 times too small. What reaches the last token depends on which token is present, not on how far the input moves. Small changes along digit directions are even carried more strongly than along word directions (3--8 times, to first order), so the digit's small trace goes with swapping one digit for another, a smaller change in kind than swapping in a word. Which mechanism carries a token's identity to the last token is open.

The same null, pre-registered for the larger models, gives the same picture in Qwen2.5-7B (word swap 0.90 and noise 0.84 of the digit's overlap; the word swap reads at 0.27--0.93, the digit at 0.13--0.27 and the noise at 0.10--0.12). In Ouro-2.6B it is mixed: the noise is as crowded as the digit (1.28 times its overlap) and as unreadable, but the word swap lands on quieter directions (0.34 times) and reads at 0.82--0.86. There, both size and crowding favour the word, and ``small, not crowded'' describes the digit against noise only.

\begin{table}[t]
\centering
\caption{Qwen2.5-1.5B, last prompt token, layers 7--28 (pre-registered, except the input distance, a descriptive check). Input distance: mean distance between the ten variants' input embeddings. Overlap / shuffled: each change's overlap over that of its own shuffled-label subspace. Signal/noise: inside the change's own 9-dimensional subspace. The same ten words and ten noise vectors are used in every family.}
\label{tab:perturb}
\small
\begin{tabular}{@{}lccccc@{}}
\toprule
change & \shortstack{input\\distance} & \shortstack{overlap /\\shuffled} & \shortstack{signal/noise\\in subspace} & \shortstack{standard\\probe} & \shortstack{minimal-pair\\probe} \\
\midrule
a different digit & 0.70 & 0.68--1.19 & 0.0009--0.0037 & 0.12--0.14 & 0.75--0.98 \\
a random word & 1.42 & 0.50--1.06 & 0.011--0.205 & 0.24--0.74 & 0.95--1.00 \\
random noise & 1.46 & 0.75--1.24 & 0.0001--0.0010 & 0.11--0.15 & 0.76--0.99 \\
\bottomrule
\end{tabular}
\end{table}

This is not superposition in the usual sense \citep{elhage2022}: the digit shares its directions with the variation between contexts, not with a fixed set of other features. Probing work usually asks whether a probe's success reflects the representation or the probe's capacity \citep{hewitt2019}; here a probe fails because a signal shared across contexts is small next to the variation the contexts add along the same directions, and only the minimal-pair contrast removes that variation. The same limit is known from population coding in neuroscience, where only noise along the signal's own direction limits a decoder \citep{morenobote2014}.

\subsection{Ouro by state, and further observations}\label{app:state}

\begin{table}[h]
\centering
\caption{Ouro, last prompt token, held-out templates. Overlap: context variance per dimension inside the digit subspace over its average per dimension; shuffled: the same for a subspace fitted with the digit labels shuffled within each family. Structured decoder: a mass-mean probe with a shrunk covariance, built from the minimal-pair digit means and the training prompts, applied to raw single prompts. Gaussian ceiling: the same decoder on samples from its own fitted model. Standard probe trained on 50\% and 100\% of the training templates. No PCA: the standard probe on all 2,048 dimensions, $C$ from $10^{-4}$ to 10. MLP: one hidden layer of 256 units on the 256 principal components.}
\label{tab:swamp}
\small
\begin{tabular}{@{}lcccccccc@{}}
\toprule
state & overlap & shuffled & \shortstack{signal/noise\\in subspace} & \shortstack{structured\\decoder} & \shortstack{Gaussian\\ceiling} & \shortstack{standard probe\\50\% / 100\%} & no PCA & MLP \\
\midrule
mid 1 & 11.7 & 10.7 & 0.005 & 0.17 & 0.75 & 0.14 / 0.15 & 0.16 & 0.11 \\
end 1 & 7.2 & 9.1 & 0.014 & 0.17 & 0.90 & 0.18 / 0.19 & 0.18 & 0.13 \\
end 2 & 6.5 & 7.9 & 0.023 & 0.21 & 0.92 & 0.21 / 0.23 & 0.24 & 0.17 \\
end 3 & 9.2 & 10.1 & 0.011 & 0.20 & 0.88 & 0.17 / 0.18 & 0.19 & 0.17 \\
end 4 & 9.4 & 11.3 & 0.010 & 0.18 & 0.85 & 0.17 / 0.19 & 0.18 & 0.12 \\
\bottomrule
\end{tabular}
\end{table}

The minimal-pair probe already reads the digit at 0.87--0.92 from a quarter of the training templates. Against their own shuffled-label baselines, the colour word and the number varied in the added sentence have overlaps of 0.53--0.83 and 0.85--1.13 of the baseline in Ouro. In Qwen both are at about their baseline, except at the final layer, where the colour becomes readable to the standard probe (0.62--0.70) while the digit stays at 0.15--0.18. In a profile over all 29 depths of Qwen at the last token, the minimal-pair probe dips in the middle of the network (0.58--0.63 at layers 18--20) before it recovers to about 0.90. The top within-family residual directions have a higher overlap than the digit's subspace (15--23 in Ouro).

\textbf{Linear response at the varied token.} A pre-registered study took 24 of the questions, one per template, with their original digit and the same prompt, and measured by finite differences how the last prompt token's state responds to small changes of the digit's input embedding (step 0.02 times the median embedding norm). The squared gain along directions in which word embeddings differ is 1.16--1.28 times that along random directions in Qwen (layers 7--28) and 1.36--1.50 in Ouro (loop ends), against the 3 that would have counted as tuned to token directions; we had predicted tuned. Along digit-difference directions it is 2.7--8.5 and 4.4--5.0 times the word-direction gain. The response is linear only locally: for the full word swap, the linear prediction is orthogonal to the actual change ($\cos \approx 0$) and 4.5--15 times too small, while the full-size noise acts roughly linearly ($\cos$ 0.38--0.68). Per unit input, a word swap changes the last token 10--25 (Qwen) and 38--373 (Ouro) times more in squared size than the noise, and a digit swap 0.12--0.94 and 0.23--0.36 times as much as a digit-to-word swap. Causal tracing corrupts inputs by adding Gaussian noise to token embeddings \citep{meng2022locating}, and \citet{zhang2024patching} recommend replacing tokens instead; here, noise of a token embedding's size barely reaches the summary position, while a token swap does.

\section{Ouro: the computation studies in detail}\label{app:ouro}

\subsection{Studies 1 and 2}\label{sec:studies}

Studies 1 and 2 used two-step word problems whose answer is $a+b-c$, so the intermediate $a+b$ is never written; for example, ``Maya has 30 stickers. Her friend gives her 30 more stickers. She then uses 13 stickers on a card. How many stickers does Maya have left?'' A study has 12 templates $\times$ 3 families $\times$ 200 of the 1,000 unit-digit combinations (7,200 problems), plus bare additions (``What is A plus B?'') with all 100 unit-digit pairs per family. Study 2 used 12 new templates with new wording, objects and names. Both passed their gates: answer-only accuracy after loop 4 was 0.66 and 0.60, the copied digits read at 0.90 or above, and on bare problems the sum's units digit read 1.00 at P1, where it is Ouro's next prediction. Intervals are 97.5\% (Bonferroni over the two hypotheses) and resample families. Table~\ref{tab:prereg} gives the pre-registered results.

\begin{table}[t]
\centering
\caption{Pre-registered results, after loop 4: excess over the copy ceiling. Study 1 was scored mod 10 against $a-b$ and $a+c$; Study 2 mod 5 against $a-b$, $a+c$ and two mixtures of the answer with $c$.}
\label{tab:prereg}
\small
\begin{tabular}{@{}L{0.26\linewidth}L{0.33\linewidth}L{0.33\linewidth}@{}}
\toprule
 & Study 1 & Study 2 (new templates) \\
\midrule
$a+b$ at P1 & +0.205 [0.172, 0.236] & +0.219 [0.194, 0.244] \\
margin over best control & +0.118 [0.094, 0.141] ($a-b$) & +0.075 [0.052, 0.098] (answer $+\,2c$) \\
bar for the margin & 0.15 & 0.025 (set after Study 1) \\
label & UNCLEAR & passes; fails Study 1's 0.15 \\
$a+b$ at P0 & $-0.016$ & $-0.007$ \\
bare addition at P0 & sum 0.71 vs ceiling 0.10 (no controls) & $a+b$ +0.372 vs $a-b$ +0.463: mixing \\
\bottomrule
\end{tabular}
\end{table}

\begin{figure}[t]
\centering
\includegraphics[width=0.82\linewidth]{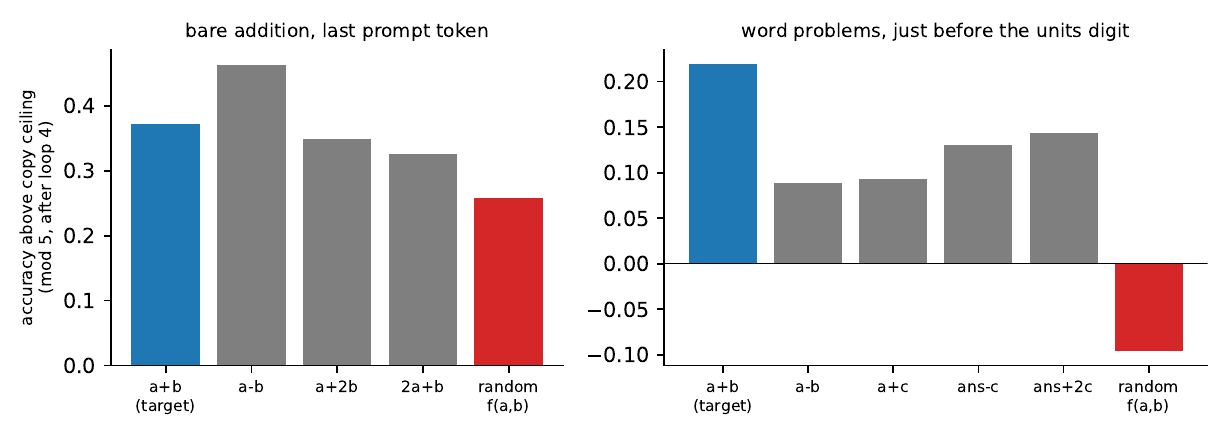}
\caption{Study 2, after loop 4, mod 5: excess of the minimal-pair probe over each label's copy ceiling. Left: bare addition at the last prompt token. Right: word problems at P1, just before the answer's units digit.}
\label{fig:comp}
\end{figure}

Study 1's controls did not include combinations of the answer digit with $c$; since $a+b = \text{answer} + c$, a state holding the answer and a copy of $c$ could make $a+b$ readable without any intermediate. Study 2 added those controls and moved to mod 5. Before the answer starts, at P0, $a+b$ is not readable in either study; Study 3 returns to P0. In bare addition, the apparent sum before writing is generic mixing: with controls, $a-b$ reads better than $a+b$ (Figure~\ref{fig:comp}, left).

\subsection{Sign flips and every combination}\label{sec:signs}

The analyses in this subsection were run after the two studies, on their saved states (Appendix~\ref{app:status}). The sharpest one compares $a+b$ with its sign-flipped twin $a-b$ (Table~\ref{tab:signs}). In both studies $a+b$ reads 0.09--0.13 better after loop 4 and 0.16--0.17 better after loop 3, on every one of the 24 templates. Here the needed sign is plus, so a general preference for differences, which is what bare addition showed, predicts the opposite. For $a-c$ against $a+c$ and $b-c$ against $b+c$ the needed sign is minus, so the same preference predicts the observed winners and the comparison cannot separate the two; $b-c$ also loses to $b+c$ after loop 3.

\begin{table}[t]
\centering
\caption{Sign-flipped twins at P1 (excess over each label's copy ceiling; difference with 97.5\% interval over families). Only the first row has a plus as the needed sign.}
\label{tab:signs}
\small
\begin{tabular}{@{}lcccc@{}}
\toprule
needed vs flipped & Study 1, loop 4 & Study 2, loop 4 & Study 1, loop 3 & Study 2, loop 3 \\
\midrule
$a+b$ vs $a-b$ & +0.087 [0.069, 0.105] & +0.130 [0.105, 0.154] & +0.173 & +0.161 \\
$a-c$ vs $a+c$ & +0.122 [0.093, 0.152] & +0.065 [0.038, 0.092] & +0.064 & +0.044 \\
$b-c$ vs $b+c$ & +0.052 [0.027, 0.079] & +0.042 [0.015, 0.069] & $-0.032$ & $-0.041$ \\
\bottomrule
\end{tabular}
\end{table}

\begin{figure}[t]
\centering
\includegraphics[width=0.9\linewidth]{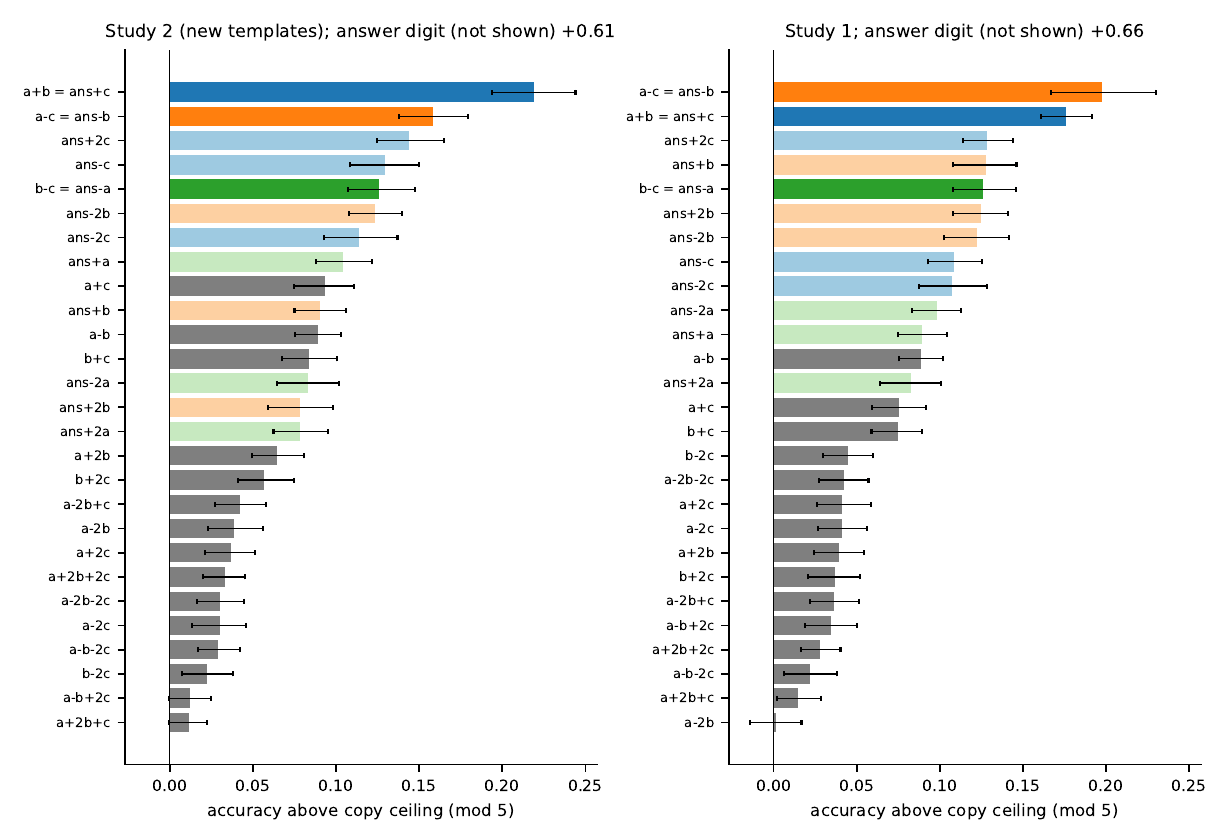}
\caption{Every mod-5 combination of the operands' units digits at P1, after loop 4: excess over its own copy ceiling, with 97.5\% intervals over families. Each colour is one line through the answer (answer $+\,k\,\times$ one copied number); the dark bar is that line's two-number member ($a+b$, $a-c$, $b-c$). Grey: combinations on no answer line. The answer digit and the copies are not shown.}
\label{fig:allforms}
\end{figure}

Against all 30 combinations of the digits other than the answer (which Ouro is about to write, and which reads +0.61 and +0.66), $a+b$ ranks first in Study 2, 0.061 [0.028, 0.093] ahead of the runner-up, and second in Study 1, 0.022 behind it (Figure~\ref{fig:allforms}). By the rule we fixed before this check, that weakens the claim that $a+b$ itself is held. The runner-up in both studies is $a-c$. It is another route to $a+b-c$, but it is also the kind of difference a general preference would favour, so we do not count it as evidence of computation.

A further test, used only in Studies 1 and 2, checks mixtures of the output. On units digits, $a+b$ is the answer plus $c$, so a state that mixes the answer it is about to write with a copy of $c$ could make $a+b$ readable. $a+b$ tops the line answer $+\,k c$ by +0.075 [0.052, 0.098] in Study 2 and +0.048 [0.030, 0.066] in Study 1, but a calibrated null on Study 3's items shows that generic mixing tops such a line too (Section~\ref{sec:early}); $a+b$ also tops the other two-number combinations by 0.09--0.13. Such mixtures do exist: after loop 4, forms on the answer lines read 0.06--0.07 above forms on no answer line. The margin also survives the stricter baselines. Adding the answer digit already written changes nothing, and adding the carry and borrow, which Ouro does not see as tokens, lowers the Study 2 margin to +0.048 [0.023, 0.074]. On problems Ouro answers correctly it is +0.064 [0.037, 0.091], and resampling templates instead of families does not widen the loop-4 intervals.

No probe label can separate ``$a+b$ was formed and the answer computed from it'' from ``the answer was formed and $c$ added back'': they are the same function of the digits. The timing does not help the first reading: after loop 1, $a+b$ and the answer digit read about equally (+0.16 each in Study 2), and after loop 2 the answer reads higher (+0.53 against +0.38). At the moment it is needed, then, $a+b$ is readable beyond what copies allow and beyond its sign-flipped twin, in two studies with different templates, and it tops its answer line by about as much as generic mixing does. Two questions remain. Every problem needed $a+b$, so a fixed preference for sums at P1 would also explain the sign-flip result. And whether Ouro uses the value needs an intervention. Study 3 addresses both.

\subsection{Study 3: swapping the operation, and patching}\label{sec:study3}

Study 3 was pre-registered in two parts: Part A, the operation swap, and Part B, patching. Its 12 new templates come in two versions that differ only in the first step's phrase: ``gave her $B$ more stickers'' gives $a+b-c$, and ``took $B$ of her stickers'' gives $a-b-c$. A plus family and its minus twin share names, tens digits and unit-digit combinations. Each version has 7,200 problems, and probes are trained and scored within a version. Let $D_+$ be the excess of $a+b$ minus that of $a-b$ on plus problems, $D_-$ the reverse on minus problems, and $I = (D_+ + D_-)/2$. A fixed preference for either sign adds the same amount to one $D$ and takes it from the other, so it cancels in $I$; we report it as $F = (D_- - D_+)/2$. The label FOLLOWS required $I \geq 0.05$ at P1 after loop 4, with its interval above zero and both $D$s positive.

All gates passed: accuracy after loop 4 was 0.67 on plus and 0.77 on minus problems, the copies at P0 read at 0.98--1.00 at their best state, and the bare sum digit at 1.00. The label is FOLLOWS (Table~\ref{tab:study3}): $I = +0.157$ [0.141, 0.172], with 97.5\% intervals that resample twin families together. On minus problems $a-b$ reads +0.27 and $a+b$ +0.08, nearly the mirror image of plus problems. The fixed preference is small and favours differences, $F = +0.032$ [0.012, 0.053], as bare addition suggested. The tens digit, fed in at P1 and written next at P0, depends on the carry or borrow from the units, which is sign-specific, so a pre-registered stricter check adds the carry and borrow to the copy ceiling. With them, $I$ at P1 is +0.126 [0.107, 0.145], still well above the bar, and at P0, before the answer starts, it is $-0.013$ [$-0.028$, 0.002]: the accuracy swap finds nothing sign-specific there beyond the carry and borrow, and the share swap is UNCLEAR there. The copy of $b$ reads only 0.77--0.81 there after loop 4 (0.66 in Study 1), so P0 is the weaker test. In both versions the needed form also tops the other forms on its answer line after loop 4, by +0.036 and +0.045, which generic mixing matches (Section~\ref{sec:early}). So the sign-flip result of Studies 1 and 2 is not a preference for sums at P1: the better-read sign is the one the problem needs. This still does not separate a first step formed before the answer from the answer with $c$ added back, since in both versions the needed form is the answer plus $c$. These are the accuracy-based swap's numbers; the share swap's re-score of P1 is in Section~\ref{sec:ouro} (FOLLOWS, $I/N = 0.63$, and 0.57 with the carry and borrow regressed, against a bar of 0.25 set on this design).

\begin{table}[t]
\centering
\caption{Study 3, operation swap (pre-registered): P1 after loop 4, excess over the copy ceiling (mod 5), with a 97.5\% interval over twin families.}
\label{tab:study3}
\small
\begin{tabular}{@{}lcc@{}}
\toprule
 & plus problems (need $a+b$) & minus problems (need $a-b$) \\
\midrule
excess of $a+b$ & +0.207 & +0.078 \\
excess of $a-b$ & +0.082 & +0.267 \\
needed minus flipped & $D_+ = +0.124$ & $D_- = +0.189$ \\
\midrule
\multicolumn{3}{@{}l}{$I = +0.157$ [0.141, 0.172]: FOLLOWS; \ fixed preference $F = +0.032$ [0.012, 0.053]} \\
\bottomrule
\end{tabular}
\end{table}

Part B asks whether Ouro uses the value. For two plus problems $x$ and $x'$ from the same family, we run both to a patch point, replace $x$'s component in a 9-dimensional subspace $U$ at P1 with $x'$'s, $h \leftarrow h + UU^\top(h_{x'} - h)$, run the remaining loops, and read the digit predicted at P1. Each subspace is spanned by the family-centred class means of one units digit ($a+b$, $a-b$ or the answer) at the patch point, fitted on the other half of the templates; a random subspace is a further control. The two problems share $c$ and have different units digits of $a+b$ with the same parity, so a parity code alone cannot produce a hit, and Ouro answers $x$ correctly. A hit is $x'$'s answer digit, the one $x$ would write if its $a+b$ were $x'$'s, and the answer patch is the positive control. The statistic is hit($a+b$) minus the larger of hit($a-b$) and hit(random): USED at 0.10 or more with the interval above zero, NOT-USED if the interval's upper end is below 0.02, and unscorable if the answer patch hits less than 30\% of the time. The frozen plan patched at the end of loop 3. After the states were extracted, but before any analysis, we added a patch point at the end of loop 2 and a second kind of pair with equal answers ($c$ shifts to compensate). For those pairs a hit is the digit $x$ would write from $x'$'s $a+b$ and its own $c$, and a state that holds only the answer with $c$ added back predicts no hits. Each of the four tests has a 99.375\% interval (Bonferroni: 2.5\% for Part A's test, 0.625\% for each patching test).

\begin{table}[t]
\centering
\caption{Study 3, patching at P1 on plus problems (pre-registered; the loop-2 point and the equal-answer pairs were added before analysis), 2,000 pairs per row. A hit is the digit the patch implies. Same-$c$ pairs move the answer with $a+b$; equal-answer pairs hold it fixed and have no answer patch. Statistic: hit($a+b$) minus the larger of hit($a-b$) and hit(random), with a 99.375\% interval.}
\label{tab:patch}
\small
\begin{tabular}{@{}llccccll@{}}
\toprule
 & & \multicolumn{4}{c}{hit rate by patched subspace} & & \\
\cmidrule(lr){3-6}
pairs & patched after & answer & $a+b$ & $a-b$ & random & statistic & label \\
\midrule
same $c$ & loop 3 & 0.902 & 0.017 & 0.032 & 0.001 & $-0.015$ [$-0.027$, $-0.004$] & NOT-USED \\
same $c$ & loop 2 & 0.331 & 0.007 & 0.006 & 0.002 & +0.001 [$-0.005$, +0.007] & NOT-USED \\
equal answer & loop 3 & -- & 0.003 & 0.002 & 0.000 & +0.001 [$-0.003$, +0.007] & NOT-USED \\
equal answer & loop 2 & -- & 0.003 & 0.003 & 0.001 & +0.001 [$-0.002$, +0.003] & NOT-USED \\
\bottomrule
\end{tabular}
\end{table}

All four tests are NOT-USED (Table~\ref{tab:patch}). At the end of loop 3 the answer patch moves the prediction to $x'$'s digit in 90\% of pairs, so a 9-dimensional patch at P1 can steer the output. The $a+b$ patch does so in 1.7\%, slightly less often than the $a-b$ patch, and under it Ouro keeps its own digit in 96--99\% of pairs. With the answer held fixed, the $a+b$ patch hits 0.3\%. The end-of-loop-2 tests are weak: there the answer patch hits only 33\%, because loops 3 and 4 restore $x$'s own digit in about two thirds of pairs, which fits a state that is re-derived in each loop \citep{wangreid2026}. Together the two parts say that at P1 the readable combination follows the operation, but the patches show no clear sign that Ouro reads it there to write the units digit (against the random subspace alone, post hoc, the loop-3 patch is $+0.017$ [0.008, 0.027], UNCLEAR). By the end of loop 3, though, the answer is strongly readable, so an intact answer could override a weak $a+b$ component even if $a+b$ had been used to form it earlier; Section~\ref{sec:early} tests this. Only P1 was patched. A pre-registered follow-up on the same problems looked earlier, at the period ending the sentence that gives $b$ (``Her friend gave her 13 more stickers.''), where a running total would sit. There the sign did not follow the operation under the accuracy-based swap, which was not re-scored ($I$ at most +0.018 at any loop end, every upper bound at most 0.036; UNCLEAR by its rule, since no bound fell below 0.02), so its patching step was not run, and the answer was not readable at the end of the sentence that gives $c$ either. A later pre-registered study re-scored that period with the share swap, with the copies of $a$ and $b$ regressed and a bar (0.35) set on 270 null worlds built on the position's own design, none of which passes it: UNCLEAR, with $N$ 0.002 and $I$ +0.000 [0.000, 0.001], the difference preferred on both kinds of problem. Planted in null worlds at the mixing found there, a running total is FOLLOWS once its $I$ reaches 0.002--0.0045, half to just over Ouro's at P1, so one as strong as P1's signal would probably have been seen; a fainter one is not excluded. Its clamp over the sentence's tail, with the copies of $a$ and $b$ projected out, is NOT-USED ($-0.001$ [$-0.002$, 0.000] against a random subspace, 2,000 pairs), but so is replacing the whole state at every tail token (0.1\%), while replacing $b$'s own token moves the answer in 91\% of pairs. The answer reads $b$ from its own token, not from the sentence's end, so this test cannot show whether a running total is formed there.

\subsection{Patching before the answer forms, and a null for mixing}\label{sec:early}

Two further pre-registered studies address the weakest points above. The first repeats the patching on the same pairs in three ways that leave no formed answer to override $a+b$ (Table~\ref{tab:early}). A single patch at the end of loop 1, where $a+b$ and the answer read about equally, cannot be scored: loops 2--4 undo any patch made there, even one of the answer (hit 0.009). Clamping, which sets the component to the partner's at the end of loops 1, 2 and 3, does not have that problem: clamping the answer moves the written digit in 91\% of pairs, while clamping $a+b$ moves it to the partner's digit in about 4\% of pairs, below the $a-b$ control and above the random one (0.3\%), and Ouro keeps its own digit in 90--97\% of pairs. Both clamp tests are NOT-USED, including the equal-answer pairs, in which an $a+b$ patch cannot carry the partner's answer. In the third test the answer's component is first set to its family mean at the end of loop 3. Ouro then writes its own digit in only 31--32\% of pairs, so the $a+b$ component left at P1 does not rebuild the answer, and the partner's does not steer the digit: every condition, random included, hits the target at about chance (0.08--0.10). This test is weaker than the clamp. It has no positive control, and with the answer removed the output is largely scrambled, so a null there may reflect a disrupted state rather than an unused value.

\begin{table}[t]
\centering
\caption{Patching $a+b$ at P1 without a formed answer to override it (pre-registered), 2,000 pairs of each kind. Statistic: hit($a+b$) minus the larger of hit($a-b$) and hit(random), with a 99.17\% interval (six tests). In the ablation test the patch is restricted to directions outside the answer subspace; that test has no positive control, and the ablation alone changes the digit in about 68\% of pairs.}
\label{tab:early}
\small
\begin{tabular}{@{}lccc@{}}
\toprule
test & answer patch hit & same-$c$ pairs & equal-answer pairs \\
\midrule
patch at end of loop 1 & 0.009 & UNSCORABLE & UNSCORABLE \\
clamp at ends of loops 1--3 & 0.914 & $-0.016$ [$-0.034$, +0.001] & +0.005 [$-0.001$, +0.012] \\
answer ablated, end of loop 3 & -- & +0.005 [$-0.011$, +0.015] & $-0.001$ [$-0.016$, +0.013] \\
\bottomrule
\end{tabular}
\end{table}

The second study calibrates the protocol's mixture tests. We built synthetic states on Study 3's items with copies of $a$, $b$ and $c$, a strong answer code and no $a\pm b$ term, plus one generic mixture of the four digits: products of random codes for every pair, or same-frequency products of periodic digit features, or a random one-layer network. The mixing strength was set so that the needed form reads about as well as in Ouro, and the paper's own probes then scored each world. Pairwise mixing tops the answer line by as much as Ouro does or more (mean margins +0.078 and +0.051 over six worlds each, against +0.036 and +0.045), so by the pre-registered rule the line test does not separate a held $a\pm b$ from mixing. The reason is that the needed form is a function of two inputs, $a$ and $b$, which pairwise mixing makes readable directly, while the other forms on its line need all three; in the Fourier worlds, answer $\pm 2c$ barely reads, as same-frequency products predict. The random network never made the needed form as readable as in Ouro, so it could not be calibrated. The same worlds give an operation-swap statistic $I$ of +0.09 on average and at most +0.14, against +0.157 in Ouro: the sign-following signal exceeds every null world, but mixing alone produces about half of it. In small transformers trained from scratch on the same task, dense checkpoints did not separate the two accounts either: the sign-specific signal never appeared while the answer was absent, and two of three architectures solved the two-digit task with no readable first partial result (Appendix~\ref{sec:v220}). These comparisons use the accuracy-based swap; the share swap's bar comes from 420 calibration worlds instead, and 420 fresh ones gave no false positive (Section~\ref{sec:control}); 270 worlds rebuilt on Study 3's design set Ouro's bar at 0.25 (Section~\ref{sec:ouro}).

\subsection{Loop dynamics}\label{sec:dynamics}

At P1, every combination of the numbers we checked reads better at the end of a loop than in its middle, including $a+c$, which the problem never needs, while the copy of $b$ reads better in the middle. This is not an effect of normalization: Ouro normalizes the state it feeds into the next loop, but dividing each mid-loop state by its own size leaves the pattern unchanged. Within each loop, the second half of the layers makes combinations of the numbers readable. Unneeded ones then fade: $a+c$ peaks after loop 2 (+0.46) and falls to +0.09 by loop 4, while $a+b$ peaks after loop 3 (+0.45) and keeps +0.22. In Study 3 the needed form peaks after loop 3 on plus problems (+0.51) and after loop 2 on minus problems (+0.42).

\subsection{Training order in small models}\label{sec:v220}

A pre-registered study asked whether a sign-specific $a \pm b$ appears before the answer during training, which would favour a held partial result over mixing. Eighteen small transformers were trained from scratch on answer-only $a \pm b - c$: a 6-layer model, a 2-layer block looped three times, and a 2-layer model of the looped one's size, on two-digit numbers and on arithmetic mod 23, three seeds each. Probes with copy ceilings and the operation swap ran every 100 (two-digit) or 50 (mod) training steps. No run showed the signal while the answer was absent. On two-digit numbers, the looped and the size-matched models reach about 99\% accuracy with no sign-specific $a \pm b$ at any position or checkpoint ($I \le 0.034$), and the 6-layer model shows it only after the answer in two of three seeds. On mod 23, all three architectures show it with or after the answer. On a fixed bar of 0.05 the signal crosses later than the answer, but much of that is a threshold effect: the answer signal peaks near 0.97 and $I$ at 0.06--0.26, and measured against each signal's own peak, the two rise within one to three checkpoints of each other in mod 23. Looping made no difference on mod 23; on two-digit numbers the comparison is undefined, since neither the looped nor the size-matched model showed the signal. These checks use the accuracy-based swap.

\section{The re-test of \citet{kudo2026} in detail}\label{app:kudo}

\citet{kudo2026} conclude that language models ``have not derived an answer at the moment when input was fed into the model'' and compute (sub-)answers while writing the chain of thought. Their evidence is a standard probe for each token position and layer, which before the chain reads the sub-answer at 50.5\% and the answer at 17.9\%, against 95--100\% inside it and a 90\% threshold, and patching, in which the input part barely affects the chain. We applied the protocol to their task in a pre-registered study: their two-step problems ($a = k \pm b$, $b = d_1 \pm d_2$, then ``$a$?''), in their format with three worked examples, in Qwen2.5-7B, which writes their chain exactly in 99.8\% of problems. We read the states at two positions before the chain, the end of $b$'s equation and the last prompt token. A family varies $d_1$ and $d_2$ over every valid pair, in a plus and a minus version; the constant $k$ varies across families, which separates $b$ from the answer. We also ran their short chain format (``$b = 5$, $a = 6$''), in which the chain does not restate $b$'s equation.

Their probe replicates: on single prompts it reads $b$ before the chain at 0.64 (their 50.5\%) and $a$ at 0.32 (their 17.9\%). The protocol shows what that readability is (Table~\ref{tab:kudo}). At both positions and in both formats the operands are combined beyond the copy ceiling: $(d_1+d_2) \bmod 5$ and $(d_1-d_2) \bmod 5$ read 0.12--0.38 above it. The original accuracy-based swap called neither position FOLLOWS ($I = 0.00$ and $+0.03$ against its bar of 0.05), but that swap later failed its positive control (Section~\ref{sec:control}). Re-scored with the validated share swap (Table~\ref{tab:kudo}), both positions are UNCLEAR: the state prefers the difference on both kinds of problem ($D_+ < 0$), with a small part that follows the operation ($I = +0.020$ and $+0.010$, intervals clear of zero). By the pre-registered rule, nothing places $b$ before the chain. Nor does the chain use such a value. In their format, replacing the whole state at either position with another problem's, from layer 12 on, never changed the $b$ the chain writes in a 40-pair check before the freeze (0 of 40; the main format's patch was not run as a pre-registered test): the chain restates $b$'s equation and computes from that text. In the short format, replacing the state at the end of $b$'s equation changes the written $b$ to the other problem's in 96\% of pairs, and so does replacing only the directions of the copied digit (96\%); the directions of $b$ with the copy projected out do so in 2\%, and the last prompt token feeds that step in none.

Nothing in the protocol's results places the sub-answer before the chain, and the chain computes it from the operands, but the swap cannot tell at these positions. They also show how two weaker tests would have disagreed (their probe, at 0.64, is below their 90\% threshold): the operands are combined well beyond the copy ceiling, and clamping $b$'s subspace without removing the copied digit moves the written $b$ in 30\% of pairs. Both suggest an early sub-answer. The share swap does not find a code that follows the operation (UNCLEAR), and the copy-controlled clamp finds $b$'s own directions barely move the written $b$, and it keeps 75--85\% of them, more than a clamp of this form kept when it detected a used value in small models (post hoc; Appendix~\ref{app:v230}). Their probe also misreads a value that is certainly present (post hoc, descriptive): on the same replication states it reads the copied operand $d_1$ at 0.59 at the last prompt token, below their 90\% threshold, though minimal pairs read it at 1.00 there. It reads the other operand, $d_2$, at 0.998 at the same token, and both at 1.00 at the end of $b$'s equation, so its rule misses one of the two operands, and only at the end of the question.

\paragraph{Without a chain of thought.} \citet{kudo2026} also have the model answer directly (``$a = 6$''). Their probe then reads $b$ at 69\% and $a$ at 41\% before the answer, both below their threshold, and they conclude that when the model fails, the sub-answers are unlikely to be represented in its states. Their probe replicates here as well: run as in the chain study (post hoc), it reads $b$ at 0.74 at the token before the answer digit (0.63 within the prompt) and $a$ at 0.32 within the prompt. In this setting the model has to form $b$, or find a shortcut, without writing it. Qwen2.5-7B answers 69\% of random problems like theirs directly (12,000 problems; their 77.8\%; our pilot, 67.5\%). The gap of about 9 points may reflect differences between our re-implementation of their format and theirs. On our families it answers 76\%: 88\% when $b$'s operation is a plus and 64\% when it is a minus, falling to 38\% when both operations subtract. In a second pre-registered study we read the same positions plus a third, the token just before the answer digit. There the answer is itself a confound, since within a family it fixes $b$. The study therefore also clamps the answer's directions, projects them out of $b$'s along with the copy's, and calibrates a null that mixes the copies with the answer and has no $b$ in it. The operands are again combined beyond the copy ceiling at every position (+0.14 to +0.38), but under the original accuracy-based swap no position followed the operation ($I$ from 0.00 to +0.04, against its bar of 0.05), and mixtures of the copies with the answer, with no $b$ in them, gave as much (up to +0.06). Re-scored with the share swap and the answer regressed out, every position is UNCLEAR (Table~\ref{tab:kudo}). The clamp gives the same picture. Replacing the whole state with another problem's changes the answer to that problem's in 85\% of pairs at the two scorable positions. At the end of $b$'s equation the copied digit's directions alone do this in 83\% of pairs. Just before the answer, the answer's own directions do it in 46\% and the copy's in 24\%. $b$'s directions with both the copy and the answer projected out do it in 1\% and 3\%; without the projection, $b$'s directions do it in 28\% and 35\%. Without a chain, then, $b$'s directions beyond the copies and the answer barely move the answer, but with 35--54\% of them kept that test has little shown power (Appendix~\ref{app:v230}); and since the answer fixes $b$ within a family, the design cannot say whether $b$ is held inside the answer's code. Here too the protocol cannot tell. Their sentence about failures is not contradicted either. Split by whether the model answers correctly, both our operation signal and their probe at first point the other way: on wrongly answered problems $I$ is larger, and their probe reads $b$ at 0.77 against 0.73. But two thirds of the wrong answers come from problems in which both operations subtract, where $b$ reads best. Compared within each pair of operations (post hoc, rule recorded before running), their probe reads $b$ slightly worse on failures ($-0.04$, 95\% CI $[-0.09, +0.01]$).

\paragraph{Equal answers (pre-registered).} The no-chain clamp had to project $b$'s subspace off the answer's, because within a family the answer fixes $b$. A third study used new families (14,400 problems) that also vary $k$: $k \in \{0, 1, 2\}$ when $a = k + b$ and $\{9, 8, 7\}$ when $a = k - b$, crossed with every $(d_1, d_2)$ with $d_2 \le d_1$ and $d_1 + d_2 \le 7$, in a plus and a minus version, so that every answer is one digit. A pair shares the family, the version and $d_1$, differs in $d_2$ (so in $b$ and $k$), and has the same answer; both problems are answered correctly. A used $b$ then predicts $P = k \circ b'$, $x$'s $k$ with $x'$'s $b$, which neither prompt has. The clamp follows the no-chain study (layers 12--24, class means fitted on the other half of the families), with $b$'s subspace projected off $d_2$'s and $k$'s copies only; this keeps 85\% of $b$'s directions. On 1,000 pairs per position, at the end of $b$'s equation the answer becomes $P$ in 0.865 of pairs with the whole state (from layer 11), 0.884 with $d_2$'s copy, 0.203 with $b$'s unprojected subspace, 0.039 with $b$'s directions beyond the copies, 0.099 with the answer's subspace (a built-in null, since the answers are equal) and 0.001 with a random one; the primary statistic is $+0.038$ [0.023, 0.056], UNCLEAR. Before the answer digit and at the last prompt token the whole state reaches $P$ in 0.000 and 0.002 of pairs (unscorable). The pre-registered design check (the answer's subspace against random) was omitted by the frozen analysis code and computed afterwards from the saved pairs as specified: $+0.098$ [0.077, 0.121], just under its 0.10 bar, so no label changes; $b$'s subspace lies about 40\% in the answer's, so clamping the answer's also swaps part of $b$. The scorability gate was moved from the copy's patch to the whole state's after a smoke run with random subspaces, before any learned subspace was used.

\paragraph{Where the answer reads its inputs (pilot).} In these equal-answer pairs the end of $b$'s equation is $d_2$'s token, which differs between the prompts. The relay study therefore added pairs that share $d_2$ and differ in $d_1$, from the same items. A pilot (300 pairs per type, both problems answered correctly, no learned subspace) replaced the whole state at one of ten positions ($d_1$, its operator, the space, $d_2$, `` ;'', the question's letter, ``?'', the target's letter, `` ='' and the token before the answer digit) after one of layers 2, 4, \dots, 16, 20 or 24, or after every layer. The answer became $P$ only when the differing operand's own token was replaced: 0.80--0.93 after layers 2--20 and 0.91 after every layer for $d_1$; 0.87--0.92 through layer 16, 0.62 after layer 20 and 0.88 after every layer for $d_2$; 0.00 after layer 24 for both. Every other position gave at most 0.06, and the token before the answer digit up to 0.19 (after layer 20, in pairs differing in $d_2$).

\paragraph{The relay test (pre-registered).} 1,000 new pairs per type (the pilot's pairs excluded in either order); the two problems of a pair also differ in $k$. The primary row replaces the five tokens strictly between $d_2$ and the token before the answer digit (five in every one of the 14,400 problems) after every layer 1--27; the gate is the same patch at the differing operand's token, which must reach 0.30. The statistic is the share of pairs whose answer becomes $P$, against no patch, with a 97.5\% cluster bootstrap over families (10,000 resamples); NOT-HELD if the upper bound is below 0.02, HELD if the share is at least 0.10 with the lower bound above 0. Results: pairs differing in $d_1$, $+0.003$ [0.000, 0.007], gate 0.898; in $d_2$, $+0.002$ [0.000, 0.006], gate 0.876; both NOT-HELD, and the answer stays $A$ in 99.4\% and 99.3\% of pairs. The other operand's token after every layer gives 0.058 and 0.013. After single layers 16, 18, 20, 21, 22, 23 and 24 the operand's token gives 0.80, 0.79, 0.79, 0.78, 0.70, 0.02, 0.00 ($d_1$) and 0.87, 0.84, 0.61, 0.60, 0.40, 0.01, 0.00 ($d_2$); the token before the answer digit gives 0.00, 0.00, 0.03, 0.03, 0.06, 0.05, 0.00 and 0.02, 0.03, 0.19, 0.20, 0.26, 0.09, 0.00. No run on the model preceded the freeze, since with whole-state patches it would have shown real outcomes; the mechanics were checked on a small random model, in which replacing every position after every layer reproduces the partner's output exactly.

\paragraph{Different answers (pre-registered) and a position map (post hoc).} The fifth study repeats the relay test with pairs that share the family, the version and $k$ and differ in $d_1$ (with $d_2$ shared) or in $d_2$ (with $d_1$ shared), so that $b$ and the answer differ; hit = the partner's answer, which a carried $b$, operand or finished answer all predict. Rows, gate, bootstrap (seed 2383) and labels are the relay test's; candidates are shuffled with seed 2380, 1,000 pairs per type, both answered correctly. Results: differing in $d_1$, $+0.081$ [0.061, 0.103], gate 0.998; in $d_2$, $+0.083$ [0.062, 0.107], gate 0.999; both UNCLEAR, and the answer stays $A$ in 86.5\% and 88.7\% of pairs. The other operand's token gives 0.069 and 0.000. After single layers the operand's token gives 0.67 and 0.40 after layer 22, 0.03 and 0.01 after 23; the answer position gives the partner's answer in 0.13 and 0.45 after layer 22, 0.76 and 0.95 after layer 23. No run on the model preceded the freeze; the design is the relay test's. A post hoc map on the same pairs (no decision rule; 97.5\% cluster intervals, seed 2393) replaced each of the five positions alone, all but one, or all five after a single layer: `` ;'' alone 0.008 and 0.014, the question's letter 0.000 and 0.003, ``?'' 0.000 and 0.001, the target's letter 0.000 and 0.000, `` ='' 0.058 [0.041, 0.077] and 0.054 [0.037, 0.074]; all but `` ='' 0.011 and 0.020, all but any other position 0.066--0.084; all five after layer 8, 12, 16, 20, 22 or 23 only, 0.001, 0.003, 0.006, 0.030, 0.024, 0.044 and 0.000, 0.003, 0.008, 0.041, 0.033, 0.042.

\section{The two-hop control in detail}\label{app:twohop}

\paragraph{Items.} A first pilot asked Qwen2.5-7B for each hop separately: country to capital 25 of 25; city to country 0.97 of 1,083 cities from Wikidata; the two-hop prompt 0.98 (0.995 where it names the country right), scored on the capital's first token, which agrees with scoring the whole continuation on 0.995; person to birth country 0.93 and the person two-hop prompt 0.86 (122 people). The study uses the 1,038 cities the model gets right on both (25 countries, 8--60 cities each; names that appear in two countries dropped) and 99 people, with three worked examples from countries outside the 25. The outcome is the argmax over the 25 capitals' first tokens, which are distinct.

\paragraph{Ground truth (second pilot, no learned subspace).} Replacing the whole state at the city's last token after one layer moves the answer to the other country's capital in 0.82--0.84 of prompts (layers 4--12), 0.72 (16), 0.58 (20) and 0.00 (24); a person's state does so in 0.63--0.70 and a country name's in 0.96--1.00 (layers 4--16), and a city of the same country keeps the answer (0.99). At the last token every source gives 0.00--0.05 through layer 20 and 1.00 at layer 24, where the second hop has been done. A nearest-class-mean readout on half the cities reads the country at the city's token at 0.32 (layer 0), 0.54 (4) and 0.94--0.96 (8--27), and at the last token at 0.03, 0.25, 0.81 (8) and 0.92--1.00 (12--27), against a copy ceiling of 0.38 from the name tokens' embeddings (chance 0.04). Projecting the country's subspace off the copy's keeps 0.99 of it. The band (layers 8, 12, 16) and the two positions were chosen from these numbers.

\paragraph{The clamp (pre-registered).} Per position, layer and half of the cities, $U$ is spanned by the 25 country class means (centred; 24 dimensions), $C$ by the same from the name tokens' mean embedding, $U_o$ is $U$ projected off $C$, and $R$ is random with $U_o$'s dimension; a pair uses the bases fitted on the half without $x$. The verdict required the whole state at the band to be USED at the city's token and NOT-USED at the last, a same-country design check to pass, and then read the clamp at both. Results on 1,038 pairs: at the city's token $U_o$ 0.789, $U$ 0.793, $C$ 0.000, $R$ 0.000, whole state 0.849 ($U_o - R$ $+0.789$ [0.732, 0.841], USED); at the last token every condition 0.000 (NOT-USED); same-country pairs change no more under $U_o$ than under $R$ ($+0.000$ [$-0.003$, +0.003]). Verdict PASSES. People's states at the city's token: $U_o$ 0.524, whole state 0.760. A smoke run with every learned subspace replaced by a random one found, before the freeze, that some last-token bases had 25 dimensions from float32 rounding; they are now fitted in float64 and capped at 24.

\section{An unwritten value at a distance in detail}\label{app:v230}

\paragraph{Task and models.} Prompts are ``[pad] $x \circ z$ ; [$k$ fillers] $w$ ='' with $x, z, w \in \{0, \dots, 22\}$, $\circ \in \{+, -\}$, $k = 6$ or $0$ (pads keep the length fixed), and answer $y = (T[(x \circ z) \bmod 23] + w) \bmod 23$ for a fixed permutation $T$. The intermediate $s = (x \circ z) \bmod 23$ is never written, and since $w$ varies the answer does not fix it. The models are 6-layer transformers ($d = 192$) trained from scratch; five seeds reach 0.995--1.000 held-out accuracy, and each run reproduced, bit for bit, the checkpoint the ground truth was measured on (checked by hash). Negative controls (five seeds, 0.983--0.997) are trained on $y = (T_1[x] + T_2[z] \circ w) \bmod 23$: the same tokens, but $x \circ z$ is never needed.

\paragraph{Ground truth.} In pilots, before any protocol statistic, distributed alignment search \citep{geiger2024finding} learned a subspace at each position and layer and swapped it between two prompts. When the swap carries $s$, the model writes $T[s'] + w$ in 0.78--0.98 of held-out pairs at ``;'' (layers 3--5) in every model, and in at most 0.07 elsewhere, except 0.13--0.18 at ``='' in one model, which is not scored there. Swapping a single operand at ``;'' succeeds in at most 0.08 of pairs, so ``;'' holds $s$ itself, not copies of the operands for later use. At layer 0, where the operands are still on their own tokens, the same search swaps each of them in 0.99--1.00 of pairs. The pre-registered rule: a position is positive if $s$ reaches 0.70 with each operand at most 0.15, negative if $s$ stays at most 0.15.

\paragraph{The protocol.} The share swap is the one used for Ouro and Kudo (copies $x$, $z$, $w$ and the answer regressed out; layers 1--5), with its bar recalibrated on this design from 270 null worlds as in Section~\ref{sec:protocol} (0.35). The clamp follows Section~\ref{sec:protocol} on 1,000 pairs that differ only in $z$ (so in $s$ and the answer, not in $x$ or $w$), with $s$'s subspace projected off $z$'s copy. It differs in how the codes are found. In a check on a known answer, before the freeze ($w$'s own code at $w$'s token), class means of $z$ carried part of $w$'s code, because $w$ is drawn for each pair, and replacing $z$'s subspace moved $w$ in 40--97\% of pairs; the class codes are therefore fitted jointly, by least squares on one-hot codes of $s$, $x$, $z$, $w$ and the operation, and each subspace keeps only the directions above a shuffled-label null. On that check the clamp then gives $+0.86$ and $+0.99$ (one model unscorable).

\paragraph{Results.} At ``;'' the clamp gives $+0.90$, $+0.67$, $+0.60$, $+0.93$, $+0.97$ over the random subspace at $k = 6$ and $+0.85$, $+0.90$, $+0.62$, $+0.91$, $+0.98$ at $k = 0$, every interval's lower end at least $+0.54$; the whole state moves the answer in 0.82--0.99 of pairs and the random subspace in at most 0.003. It is USED nowhere else ($w$'s token at most $+0.09$) and in no negative control. The share swap gives $I/N$ 0.78--0.91 at ``;'' and is FOLLOWS at all 40 scored cells of the five models, fillers, $w$'s token and ``='' included, and at none in the negative controls (largest $I/N$ 0.04). One of 270 fresh null worlds is FOLLOWS: a periodic mixture at the weakest mixing strength, $I/N = 0.363$ against 0.35, with a needed share $N = 0.0009$, about fifty times below that of any cell in the trained models. A floor on $N$ would have excluded it, but none was pre-registered, so the swap is not specific by its rule.

\paragraph{Does the earlier clamp transfer?} The clamps in Sections~\ref{sec:ouro} and~\ref{sec:kudo} used separate class means. They agree with joint codes when the other inputs are balanced within each class; none of the item sets is exactly balanced (Ouro's by sampling, Kudo's by construction, since $b$ is a function of $d_1$, $d_2$ and the operation). Run on the same five models and pairs with separate class means (post hoc), the clamp is USED at ``;'' in all five at both distances ($+0.49$ to $+0.89$ with the copy projected out, $+0.75$ to $+0.93$ without it, the form used for Ouro), and USED at no position where $s$ is absent, though also at ``='' in the one model where the search found $s$ weakly there ($+0.21$ and $+0.24$; joint codes $+0.08$ and $+0.09$). There the class means of $z$'s copy also carry part of $s$'s code, which costs the projected clamp some power.

\paragraph{Is $s$ present where the share swap fires?} The swap's FOLLOWS at the fillers, $w$ and ``='', where DAS and the clamp find no use, could be false positives or could mark a value that is present. An independent check (post hoc) decodes $s$ with a ridge classifier on held-out families, from family-demeaned states with one-hot copies of $x$, $z$, $w$, the operation and the answer regressed out on the training folds (chance 0.04). At the fillers, before $w$ is seen, $s$ reads at 0.90--1.00 in four models and 0.63--0.65 in the fifth, against 0.20--0.31 for random functions of $(x, z)$; in the negative models $s$ reads no better than random functions anywhere (0.07--0.28 against 0.08--0.30). At $w$ and ``='', $s$ reads at 0.85--1.00 in four models, but there $s$ is a function of the answer and $w$, which regressing their one-hot copies does not remove, so these positions are weaker evidence. The value is carried to the fillers and is readable there, but not used: the swap's FOLLOWS there marks presence. The 1 null world in 270 is a true false positive.

\paragraph{Power at a faint signal.} In the same five models a further pre-registered study left the forward pass untouched and degraded only the states the clamp's subspaces are fitted on, adding variance drawn once per item until $s$'s share of the remaining variance (the share swap's $N$, as a population estimate) fell from 0.32--0.47 to 0.1, 0.03, 0.01, 0.005 and 0.002; Ouro's $a+b$ is about 0.008--0.010 on the same scale. The added variance had the covariance of the model's own residual (which crowds $s$'s span at 2.5--5.5 times the average, against 6.5--11.7 for Ouro's context in the digit's span) or was isotropic, and the subspaces were fitted on all of the fitting half or on 34 items per class, Ouro's ratio of items to dimensions. With separate class means and no copy projection, the form used for Ouro, and $s$'s subspace scored against a random one, the clamp is USED in 5 of 5 models at every dose down to 0.005 in Ouro's regime ($+0.73$ to $+0.90$; isotropic, $+0.40$ to $+0.60$), and in 3 of 5 at 0.002 (the other two voided by a calibration miss). Ouro's own comparison, against the subspace of the other sign, is weak here at any dose ($+0.00$ to $+0.15$ with the model's own covariance), because that subspace moves the answer about as often. The joint-code form of the main test keeps 0--1 directions once the signal is faint or the sample Ouro-sized, and then has no effect. The study degrades the measurement, not the model: a value the model itself codes faintly was not tested.

\paragraph{Power of the copy-projected clamp.} The form used for \citet{kudo2026} also projects $b$'s subspace off the class means of the copy, and here the class means of the operands carry much of $s$'s code. A further pre-registered study repeated the degraded-fitting design with this projection, off the class means of $z$ (the procedure used for Kudo, which keeps 55--67\% of $s$'s directions) or of $z$ and $x$ together (the primary arm, since on the one model checked before the freeze it kept 51\%, inside the 35--64\% then reported for the Kudo clamps; on all five, 26--44\%), with 34 items per class, Kudo's ratio of items to dimensions. Projected off $z$, the clamp is USED in 5 of 5 models at every dose down to 0.005 ($+0.23$ to $+0.78$). Projected off $z$ and $x$, it is USED in only 3 of 5 even at full strength ($+0.26$, $+0.10$, $+0.12$); in a fourth it is NOT-USED by the rule ($-0.04$ [$-0.06$, $-0.02$]), though the clamp without the projection moves the answer in 97\% of pairs there, and at 0.005 it is USED in none. By the pre-registered reading the clamp of the Kudo re-test has no shown power at its signal level (UNINFORMATIVE): a projection that keeps about a third of a value's directions can call a used value NOT-USED. A later check (post hoc) found that the range 35--64\% mixed two definitions of the share kept (the chain study averaged over the copy's four principal angles, not over $b$'s nine directions); on the definition used here the Kudo clamps keep 75--85\% with a chain and 35--54\% without. A projection off a random subspace of the same size keeps 99.6--99.9\% of $b$, so the loss reflects $b$'s code lying in the copy's and the answer's directions. The no-chain clamps therefore sit where the primary arm had no power, and the chain clamps keep more of $b$ than the secondary arm, which had power down to 0.005.

\paragraph{Does a computed value show the probe gap?} The gap between a standard probe and minimal pairs (Section~\ref{sec:presence}) was shown for copied digits. In the same five models (post hoc), with new families that fix $x$, $w$ and the operation and vary $z$ over all 23 values, the standard probe reads $s$ at ``;'' (layer 4) at 0.99--1.00, as well as minimal pairs do, and at the middle filler at 0.86--0.99 against 0.94--0.99. These prompts have no varied context, so there is nothing to swamp a small change; the result says that the gap is not specific to copies, not what a computed value in a language model would show.

\section{Status of each analysis}\label{app:status}

``Pre-registered'': thresholds and labels were frozen in a commit before the run. ``Rule before running'': a post-hoc analysis whose reading rule was written down before its result was seen. ``Exploratory'': decision rule fixed in code before the data, without a frozen pre-registration. ``Pilot'': run before a pre-registration to choose its design, with no decision rule.

{\small
\begin{longtable}{@{}L{0.33\linewidth}L{0.06\linewidth}L{0.25\linewidth}L{0.28\linewidth}@{}}
\toprule
Analysis & Sec. & Status & Outcome \\
\midrule
\endhead
Answer-only accuracy by loop (GSM8K; Huginn check) & \ref{app:acc} & Pre-registered & passed: +16.5 points over four loops \\
Re-score with the final scorer & \ref{app:acc} & Re-run, no test & reproduces the numbers; loop 2--4 gain +3.0 [$-1.5$, +7.5] \\
Gain on GSM-Symbolic variants & \ref{app:acc} & Pre-registered & UNCLEAR: +6.4 points, below the 8.25 bar \\
Intermediate value at the last prompt token & \ref{sec:mp} & Pre-registered & unscorable: sanity check at the majority rate \\
Copied digits by position; bare arithmetic & \ref{sec:mp} & Exploratory & digit read only on its own token; bare operands 0.96--1.00 \\
Minimal pairs; irrelevant number; colour word; value check & \ref{sec:mp} & Exploratory, rule in code before data & present at 0.92--1.00 in a shared code; not a magnitude \\
Standard probe without PCA & \ref{sec:many} & Post hoc, rule before running & still 0.16--0.24 \\
Overlap, structured decoder, learning curve & \ref{sec:many}, \ref{sec:swamp} & Post hoc, rule before running & swamped: decoder at most 0.21; +0.015 from half to all templates \\
Crowding by token type; regression on the other numbers' digits & \ref{sec:swamp} & Post hoc, rule before running & colour crowded too (ratio 0.55: mixed); the other numbers' digits explain none of the unreadability (reported only here); read against own nulls below \\
Minimal pairs and swamping in Qwen2.5-1.5B & \ref{sec:many} & Pre-registered & replicates: minimal-pair 0.88, standard at most 0.14, decoder at most 0.14 \\
Overlap against its own nulls; MLP probe (both models) & \ref{sec:many}, \ref{sec:swamp} & Pre-registered (Ouro nulls known before) & overlap 0.83 and 1.01 of the shuffled-label null; MLP at most 0.17 \\
Random-perturbation null (Qwen) & \ref{sec:swamp} & Pre-registered & generic: word swap 0.89, embedding noise 1.07 of the digit's overlap \\
Input distance of digits, words and noise (both models) & \ref{sec:swamp} & Descriptive, no test & digits about half as far apart as random words; word and noise variants equally far apart (Qwen) \\
Linear response at the varied token (both models) & \ref{sec:swamp}, \ref{app:state} & Pre-registered & not tuned to token directions (gain ratio 1.2--1.5; predicted tuned); word swap's effect nonlinear; removing the most amplified directions does not help a probe \\
Many distinct questions (Qwen, up to 5,249) & \ref{sec:many} & Pre-registered & partial: best probe 0.19; minimal pairs 0.91 \\
Many distinct questions (Ouro, up to 5,249) & \ref{sec:many} & Pre-registered & climbs: best probe 0.36, +0.060 over the last doubling; minimal pairs 0.96 \\
Minimal pairs, swamping and the perturbation null in Qwen2.5-7B and Ouro-2.6B & \ref{sec:many}, \ref{sec:swamp} & Pre-registered & Qwen replicates (0.85; standard at most 0.27), null generic; Ouro mixed (0.79, below the 0.80 bar), null mixed (word swap quieter) \\
Many distinct questions to 45,249 (Qwen2.5-7B, Ouro-2.6B) & \ref{sec:many} & Pre-registered; amended before any fit (mean gain added); run split across kernels & Qwen partial: 0.52 at the last layer; Ouro shift-limited: 0.31; minimal pairs 0.86 and 0.79 \\
One reference variant instead of ten (four models) & \ref{sec:mp} & Pre-registered & partial in all four: 0.65--0.90 against 0.79--0.96 \\
Computation, Study 1 & \ref{sec:studies} & Pre-registered & UNCLEAR: +0.118 against a bar of 0.15 \\
Computation, Study 2 & \ref{sec:studies} & Pre-registered; bar set after Study 1 & word problems pass the lowered bar (0.025), not Study 1's 0.15 (mod 10 with the new controls +0.052); bare addition fails \\
Mod-5 re-scoring of Study 1 & \ref{sec:signs} & Post hoc, no rule & margin +0.048 [0.030, 0.066] \\
Sign-flipped twins & \ref{sec:signs} & Post hoc, no rule & $a+b$ beats $a-b$ by 0.09 and 0.13 \\
Every combination (31 forms) & \ref{sec:signs} & Post hoc, rule before running & $a+b$ first and second: claim weakened \\
$a-c$ as a partial result & \ref{sec:signs} & Post hoc, after the above & not separable from a preference for differences \\
Target against its answer line & \ref{sec:signs} & Post hoc, rule after the estimates were known & 6 of 6 lines \\
Target against other two-number forms & \ref{sec:signs} & Post hoc, no rule & $a+b$ leads; $b-c$ not after loop 3 \\
Shown digit and carry-and-borrow baselines & \ref{sec:signs} & Post hoc, no rule & margin stays above zero (+0.048) \\
Template bootstrap; correct-only margins & \ref{sec:signs} & Post hoc, no rule & loop-4 intervals not widened; correct-only +0.064 \\
Operation swap (Study 3, Part A) & \ref{sec:study3} & Pre-registered & FOLLOWS: $I$ +0.157 [0.141, 0.172] (accuracy swap; see the re-score row) \\
Patching at P1 after loop 3, same-$c$ pairs & \ref{sec:study3} & Pre-registered & NOT-USED; answer patch 0.90 \\
Patching after loop 2; equal-answer pairs & \ref{sec:study3} & Pre-registered amendment, after extraction, before analysis & NOT-USED in all three; loop-2 tests weak (answer patch 0.33) \\
Running total at the end of $b$'s sentence & \ref{sec:study3} & Pre-registered & UNCLEAR (accuracy swap, not re-scored): $I$ at most +0.018; patching not run \\
Copies at P0 by state, Study 3 & \ref{sec:study3} & Descriptive, rule before running & $b$ 0.77--0.81 after loop 4 \\
Re-test of \citet{kudo2026}: minimal pairs, operation swap, mixing null (Qwen2.5-7B, their Level 3, two chain formats) & \ref{sec:kudo} & Pre-registered & accuracy swap, superseded by the re-score row: combined everywhere (+0.12 to +0.38); does not follow at the end of $b$'s equation ($I$ 0.00, mixing); unclear at the last token ($I$ +0.03, beyond mixing, below the 0.05 bar) \\
Clamp before the chain (short format) & \ref{sec:kudo} & Pre-registered; amended before analysis (copy projected out) & descriptive: written $b$ follows the copied digit (96\%), $b$ beyond the copy 2\%; last token unscorable; their format: no pre-chain state feeds the step (0/40) \\
Their standard probe on our replication set & \ref{sec:kudo} & Pre-registered, descriptive & $b$ 0.64 before the chain (theirs 50.5\%), $a$ 0.32 (17.9\%) \\
Their standard probe without a chain & \ref{sec:kudo} & Post hoc, descriptive; rule from the chain study, recorded before running & $b$ 0.74 before the answer digit (theirs 69.1\%), 0.63 within the prompt; $a$ 0.32 within the prompt (40.8\%) \\
Their probe on wrongly vs correctly answered problems, within operation pairs & \ref{sec:kudo} & Post hoc, rule recorded before running & raw split favours failures (0.77 vs 0.73) through the operation mix; within pairs $-0.039$ $[-0.087, +0.008]$ (unclear), end of $b$'s equation $-0.033$ $[-0.067, -0.000]$ \\
Re-test without a chain: minimal pairs, operation swap, answer-mixing null & \ref{sec:kudo} & Pre-registered; the frozen code applied the chain study's accuracy gate (0.90) instead of the registered 0.50, corrected after the first run with the statistics unchanged & accuracy swap, superseded by the re-score row: combined at all three positions (+0.14 to +0.38); does not follow at the end of $b$'s equation ($I$ 0.00); unclear before the answer (+0.018) and at the last token (+0.035); answer-mixing null not separable at all three (null up to +0.018, +0.047, +0.057); secondary split: larger $I$ on wrongly answered problems (no label; confounded with the operations) \\
Clamp without a chain & \ref{sec:kudo} & Pre-registered; weak (35--54\% of $b$'s directions left) & end of $b$'s equation NOT-USED (+0.005; copy 0.83); before the answer UNCLEAR (+0.025 [0.014, 0.039]; answer 0.46); last token unscorable \\
Study 3 with carry and borrow in the ceiling & \ref{sec:study3} & Pre-registered & P0 explained ($-0.013$; without them +0.017 [0.007, 0.028]); P1 +0.126 \\
Patching at the end of loop 1; clamping through loops 1--3; patching with the answer ablated & \ref{sec:early} & Pre-registered & loop 1 unscorable (answer patch 0.009); clamp NOT-USED in both pair types (answer clamp 0.914); ablation NOT-USED but without a positive control (output largely scrambled) \\
Synthetic null for the mixture tests & \ref{sec:early} & Pre-registered & line test NOT-SEPARABLE (null mean margins +0.051, +0.078); null $I$ +0.09, at most +0.14 \\
Positive control: the protocol where Qwen2.5-7B writes $b$ (short chain; eq, last, and one, two and three tokens before $b$'s digit) & \ref{sec:control} & Pre-registered; code unchanged & fails on the swap: before the digit COMBINED (+0.253), swap UNCLEAR (+0.017), BEYOND-MIXING by 0.0004 (largest null from a mixture that could not be weakened to the observed excess), clamp USED (+0.607 [0.565, 0.646]); swap UNCLEAR at every chain position; clamp unscorable at ``='' (whole state 0.10); end of $b$'s equation reproduces the chain study \\
Their standard probe on the copied operands (replication set) & \ref{app:kudo} & Post hoc, descriptive & $d_1$ 0.59 at the last prompt token (below their 90\%), 0.94--1.00 earlier; $d_2$ at least 0.998 everywhere \\
Positive control at a distance: clamp at $b$'s written digit, outcome $a$ four tokens later & \ref{sec:control} & Pre-registered; readings amended after launch, before any output (written-value limit) & USED: +0.806 [0.765, 0.844]; whole state 0.89 from layer 12, 0.90 at every layer; $a = 0 + b$ 0.63 \\
Why the accuracy swap failed (positive-control states) & \ref{sec:control} & Exploratory & random functions of the operand pair read 0.99--1.00 where $b$ is about to be written; residual-variance shares still follow the operation \\
Share-based swap: bar from 420 null worlds; fresh controls & \ref{sec:control} & Pre-registered (bar 0.30 set before the controls) & Qwen2.5-7B $a$ step FOLLOWS ($I/N$ 0.65); Qwen2.5-1.5B $b$ step unscorable (accuracy 0.78 $<$ 0.80); fresh null seeds 0 of 420 FOLLOWS: PARTIAL \\
Replacement control (Qwen2.5-3B dropped at its accuracy pilot) & \ref{sec:control} & Pre-registered after the first controls' results, statistic frozen & Mistral-7B $b$ step FOLLOWS ($I/N$ 0.48): share swap VALID \\
Re-score of the targets with the share swap & \ref{sec:ouro}, \ref{sec:kudo} & Pre-registered after validation; old results known; short-chain positions seen during development & Ouro P1 FOLLOWS ($I/N$ 0.63, small share), P0 UNCLEAR; Kudo et al.'s format and answer-only: UNCLEAR everywhere, difference preferred on both kinds of problem \\
Ouro P1 with the carry and borrow regressed; null on Study 3's design & \ref{sec:ouro} & Pre-registered after the re-score (draft review) & FOLLOWS ($I/N$ 0.57; $D_+$ +0.001, interval just above zero); null 0 of 270, bar 0.25: STANDS \\
Ouro P1's FOLLOWS against the bar's multiplier and a floor on $N$ & \ref{sec:ouro} & Post hoc, on saved results & holds for multipliers up to about 3.8 and floors up to 0.005; rests on $D_+$, whose interval a 99\% level would not clear \\
Unwritten value at a distance: small transformers trained from scratch (5 seeds; negative controls; ground truth from distributed alignment search) & \ref{sec:control} & Pre-registered (ground truth and bars from pilots, before any protocol statistic) & clamp USED at ``;'' 5 of 5 at $k = 6$ and $k = 0$ (+0.60 to +0.98), none in negative controls or where the value is not held; share swap FOLLOWS 5 of 5 ($I/N$ 0.78--0.91) but NOT-SPECIFIC (1 of 270 null worlds, $I/N$ 0.36, needed share 0.0009), and FOLLOWS at every position \\
Class-mean clamp on the same models & \ref{sec:control} & Post hoc & USED at ``;'' 5 of 5 at both distances (+0.49 to +0.89) \\
Clamp power at Ouro's signal level (fitting data degraded; same models) & \ref{sec:ouro}, \ref{app:v230} & Pre-registered & POWER-AT-OURO-LEVEL: class-mean clamp USED 5 of 5 down to share 0.005 (Ouro about 0.008--0.010); joint-code form fails \\
Clamp power for the form used for Kudo (copy projected; same models) & \ref{sec:kudo}, \ref{app:v230} & Pre-registered & UNINFORMATIVE: off $z$ and $x$ (26--44\% of $s$ kept, primary) USED 3 of 5 at full strength, NOT-USED in one; off $z$ alone (55--67\% kept) USED 5 of 5 down to share 0.005; on the same definition the Kudo clamps keep 75--85\% with a chain, 35--54\% without (post hoc correction) \\
Computed value: standard probe vs minimal pairs (same models) & \ref{sec:presence}, \ref{app:v230} & Post hoc, after the fourth draft reviews & $s$ at ``;'' 0.99--1.00 for both; no gap without varied context \\
Minimal-pair readout with one sibling prompt (Ouro-1.4B) & \ref{sec:mp} & Post hoc, after the fourth draft reviews & needed digit at the last prompt token 0.79--0.84 (two siblings 0.87--0.91; all nine 0.92--0.97) \\
Ouro's patches and clamps against the random subspace alone & \ref{sec:ouro} & Post hoc, after the power study & loop-3 patch $+0.017$ [0.008, 0.027] and clamp $+0.039$ [0.027, 0.051]: UNCLEAR; the other six NOT-USED \\
Share swap at the end of $b$'s sentence, before $c$; null on its design & \ref{sec:ouro}, \ref{sec:study3} & Pre-registered after two draft reviews; the accuracy swap's result there known & UNCLEAR ($N$ 0.002, $I$ +0.000 [0.000, 0.001]); bar 0.35, no null passes; a planted $I$ of 0.002--0.0045 is detected; tail clamp NOT-USED, but the whole-tail patch also hits 0.1\% ($b$'s token 0.91) \\
Training order in small transformers (18 runs, 3 architectures, two-digit and mod 23) & \ref{sec:v220} & Pre-registered; timing check against each signal's peak post hoc & no run AHEAD; two-digit looped and size-matched NEVER ($I \le 0.034$), 6-layer with or after (2/3); mod 23 with or after (all); no looping effect \\
Two-hop recall: hop accuracy, whole-state patches, readout (Qwen2.5-7B) & \ref{app:twohop} & Pilot & country used at the city's token (0.82--0.84), not at the last token through layer 20, where it reads 0.92--1.00 from layer 12 \\
Two-hop clamp: pretrained positive control & \ref{sec:control}, \ref{app:twohop} & Pre-registered (band and positions from the pilots) & PASSES: city's token USED (+0.789 [0.732, 0.841]), last token NOT-USED (0.000); design check ok; people +0.524 \\
Clamp without a chain, equal answers & \ref{sec:kudo}, \ref{app:kudo} & Pre-registered; gate moved to the whole state after a random-subspace smoke; design check omitted by the frozen code, computed afterwards as specified & end of $b$'s equation UNCLEAR (+0.038 [0.023, 0.056]; $d_2$'s copy 0.88, whole 0.87); other two positions unscorable; design check +0.098 (below the 0.10 at which the design would fail) \\
Whole-state patch map without a chain (pilot) & \ref{sec:kudo}, \ref{app:kudo} & Pilot & answer moves mainly from the differing operand's token ($d_1$ 0.80--0.93 through layer 20; $d_2$ 0.87--0.92 through layer 16); at most 0.02 at the five positions after $b$'s equation, 0.06 at its end, up to 0.19 at the answer position \\
Relay test without a chain & \ref{sec:kudo}, \ref{app:kudo} & Pre-registered; designed after its pilot & NOT-HELD for both pair types (0.003 [0.000, 0.007], 0.002 [0.000, 0.006]); gate 0.90 and 0.88; operand tokens stop mattering after layer 23; equal answers, so a carried answer is not tested (see the different-answer row) \\
Relay test with different answers & \ref{sec:kudo}, \ref{app:kudo} & Pre-registered, after a draft review & UNCLEAR for both pair types (0.081 [0.061, 0.103], 0.083 [0.062, 0.107]); gate 0.998 and 0.999 \\
Position map of the different-answer relay & \ref{sec:kudo}, \ref{app:kudo} & Post hoc, no rule & mostly at the target line's `` ='' (alone 0.058 and 0.054; all but it 0.011 and 0.020); five-position patches matter only from layer 20 among layers 8--23; each question token at most 0.014 \\
Normalizing the mid-loop states &\ref{sec:dynamics} & Post hoc, rule before running & alternation real, but general \\
Profiles across loops; timing & \ref{sec:signs}, \ref{sec:dynamics} & Descriptive, no test & answer readable as early as $a+b$ \\
\bottomrule
\end{longtable}
}

\end{document}